\documentclass[sigconf,nonacm,pbalance]{acmart}

\usepackage{amsmath}

\usepackage{algorithm}
\usepackage{algorithmic}

\usepackage{graphicx}

\usepackage{caption}
\usepackage{subcaption}

\usepackage{booktabs}
\usepackage{multirow}
\usepackage{tabularx}

\usepackage{enumitem}
\setlist[itemize,1]{label=$\circ$, left=0.3em}
\setlist[enumerate,1]{left=0.3em}

\usepackage{wrapfig}
\usepackage[dvipsnames]{xcolor}

\usepackage[capitalize,noabbrev]{cleveref}

\theoremstyle{plain}

\theoremstyle{definition}

\theoremstyle{remark}

\newcommand{\model}{\textsc{UdonCare}}

\AtBeginDocument{%
  }

\setcopyright{cc}
\setcctype[4.0]{by}
\copyrightyear{2026}
\acmYear{2026}
\acmDOI{}
\acmConference[]{}{}{}
\acmBooktitle{}
\acmISBN{}
\begin{document}

\title{Discovering Hierarchy-Grounded Domains with Adaptive Granularity for Clinical Domain Generalization}

\author{Pengfei Hu}
\affiliation{%
  \department{School of Computing}
  \institution{Stevens Institute of Technology}
  \city{Hoboken}
  \state{New Jersey}
  \country{United States}
}
\email{phu9@stevens.edu}

\author{Xiaoxue Han}
\affiliation{%
  \department{School of Computing}
  \institution{Stevens Institute of Technology}
  \city{Hoboken}
  \state{New Jersey}
  \country{United States}
}
\email{xhan26@stevens.edu}

\author{Fei Wang}
\affiliation{%
  \department{Weill Cornell Medicine}
  \institution{Cornell University}
  \city{New York}
  \state{New York}
  \country{United States}
}
\email{few2001@med.cornell.edu}

\author{Yue Ning}
\authornote{Corresponding author.}
\affiliation{%
  \department{School of Computing}
  \institution{Stevens Institute of Technology}
  \city{Hoboken}
  \state{New Jersey}
  \country{United States}
}
\email{yue.ning@stevens.edu}

\begin{abstract}
Domain generalization has become a critical challenge in predictive healthcare, where different patient groups exhibit shifting data distributions that degrade model performance.
Still, regular domain generalization approaches often struggle in clinical settings due to (1) the absence of domain labels and (2) the lack of clinical insight integration.
To address these challenges in healthcare, we aim to explore how medical ontologies can be used to discover dynamic yet hierarchy-grounded patient domains, a partitioning strategy that remains under-explored in prior work.
Hence, we introduce \model{}, a hierarchy-pruning method that iteratively divides patients into latent domains and retrieves domain-invariant (label) information from patient data.
On public datasets (MIMIC-III, MIMIC-IV, and eICU), \model{} shows superiority over eight generalization baselines across four representative clinical prediction tasks with substantial domain gaps, highlighting the potential of medical knowledge for enhancing model generalization.
\end{abstract}



\maketitle

\section{Introduction}

The digitization of clinical data, notably electronic health records (EHR), has transformed healthcare by enabling efficient computational analysis.
Current deep learning techniques have also achieved significant gains in diagnosis, mortality, and readmission prediction tasks~\citep{poulain2024graph, jiang2024graphcare, yang2024mplite, zhu2026medeyes}.
Still, these models trained on a set of training samples (source) often suffer performance drops when applied to a new test set (target) in a different domain. Domain shift denotes distributional changes between patient groups, such as data from different hospitals or time periods~\citep{perone2019unsupervised, koh2021wilds}.
Consequently, handling domain shift is a prerequisite for alleviating performance degradation in clinical predictive models~\citep{yang2023manydg,wu2023iterative}.
It also aligns with the objective of most domain generalization (DG) methods, such as meta-learning~\citep{balaji2018metareg, dou2019domain}, feature alignment~\citep{ganin2016domain, li2018deep}, and latent-domain techniques~\citep{matsuura2020domain, wu2023iterative} for robust prediction.

In this work, we study DG problems in healthcare applications, which pose unique challenges beyond conventional DG scenarios:
(1) Domain IDs (labels) are usually unseen in EHR datasets, but most DG solutions require the presence of domain IDs~\citep{wu2023iterative}.
Thus, some studies treat each patient as a unique domain~\citep{dou2019domain,yang2023manydg}, which is overly fine-grained and unstable for model training, and others rely on broader partitions (e.g. institute \& admission year), which overlook clinical heterogeneity~\citep{zhang2021empirical, guo2022evaluation}.
(2) Even though some DG methods do not rely on domain IDs~\citep{arjovsky2019invariant, liu2021learning}, they  overlook clinical semantics.
For instance, prior DG methods~\cite{matsuura2020domain, wu2023iterative} cluster instance features to form latent domains, but the resulting partitions are highly sensitive to training data.
In practice, patient groups vary significantly, especially over longer admission periods. Existing DG methods can capture feature-level similarity, but not the progression of medical concepts.
Hence, it is crucial to construct robust domains with explicit definitions grounded in clinical relevance.

To address these challenges, we explore the following research question:
\textit{Instead of assuming the presence of domain IDs, can we leverage medical knowledge to guide models in discovering domains that are both adaptive and clinically meaningful?}
In hospitals, visiting patients are treated based on their medical history, which includes medical concepts such as diseases or medications.
For instance, when dealing with heart failure patients, hospitals may categorize heart failure as a distinct domain or group it with other cardiovascular diseases.
Similarly, heart failure corresponds to a leaf node under a higher-level node grouping cardiovascular diseases in ICD-9 ontologies~\citep{world1988international}, which motivates us to use a pruning algorithm to identify appropriate ancestor nodes for domain partitions.

\begin{figure}[t!]
    \centering
    \includegraphics[width=\linewidth]{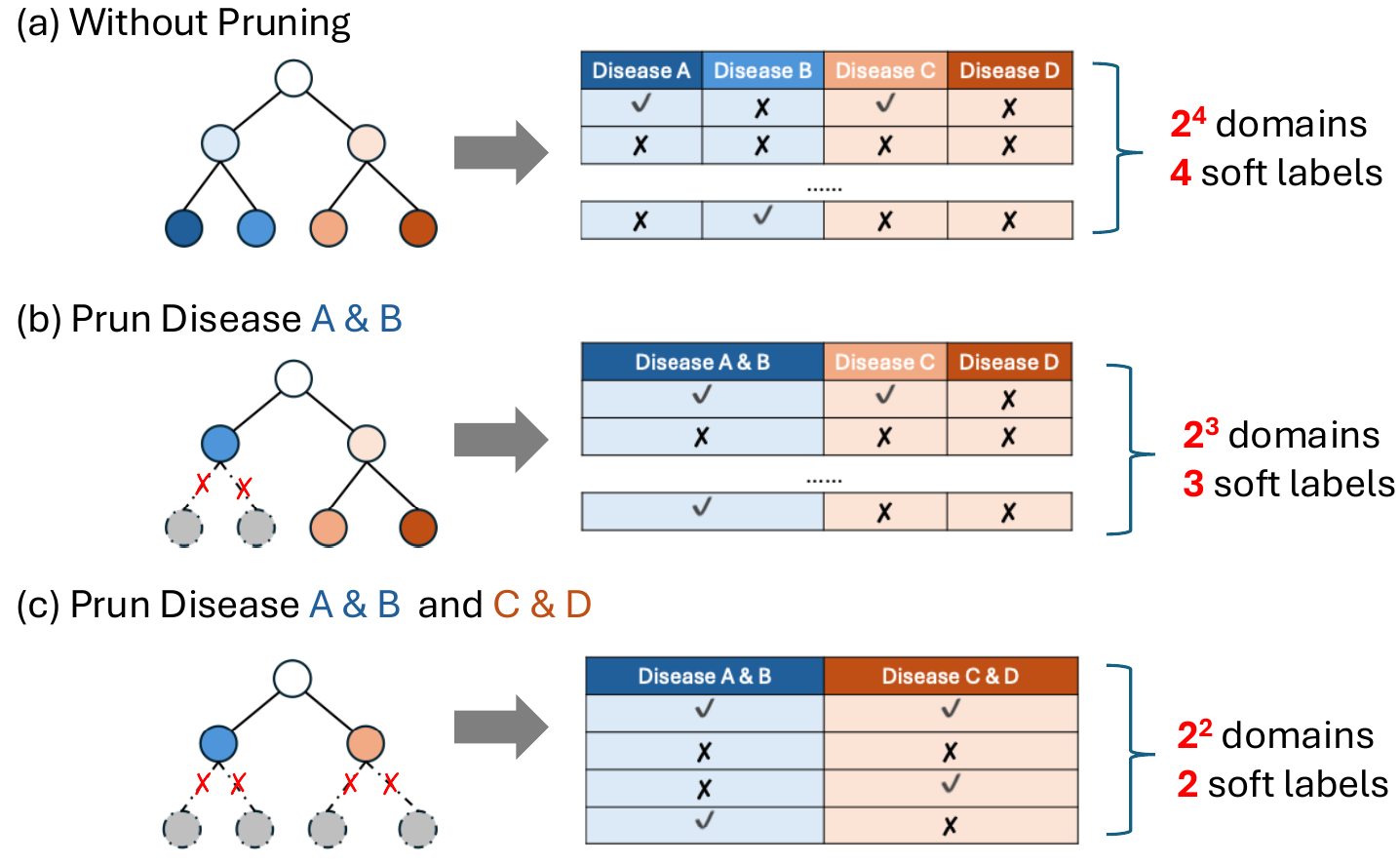}
    \caption{\textbf{An example showing how medical knowledge provides guidance to domain discovery.} We adopt the ICD-9-CM hierarchy with 4 diseases (left) to illustrate how higher-level entities can manage the possible size of generated latent domain.
    Note that, discovering fewer domains does not necessarily lead to better predictive performance.
    We present a flexible solution to discover dynamic domain labels across different tasks.}
    \Description{Illustration of four ICD-9-CM disease nodes grouped at different hierarchy levels to form candidate latent domains.}
    \label{fig:intro}
\end{figure}

However, translating medical ontologies into a form suitable for latent domain discovery is not a simple problem.
Existing DG methods identify latent domains by clustering patient-level representations in vector spaces~\citep{matsuura2020domain, wu2023iterative}, which precludes the direct incorporation of ontology hierarchies.
Meanwhile, medical ontologies are commonly used to enrich patient representations via graph neural networks~\citep{jiang2024graphcare}, but these ontologies are rarely explored as structural rules for discovering latent domains.
Thus, we aim to bridge this disconnect and leverage ontologies not as auxiliary features, but as principled constraints that shape domain formation in a clinically meaningful way.

To this end, we introduce \model{}, a framework that discovers \textbf{U}nseen \textbf{do}mai\textbf{n}s with adaptive granularity to enhance model generalization in predictive health\textbf{care}.
It ensures domain partitions remain consistent with clinical semantics while maintaining flexibility for adaptive generalization (see Figure~\ref{fig:intro}).
Concretely, a pruning algorithm is developed to merge similar concepts on hierarchies and generate soft labels as domain IDs for patients.
To explicitly remove covariates from patients, we propose a mutual learning network, which only retrieves label information upon the orthogonal factorization for final predictions.
During training, model parameters are jointly optimized in an iterative manner across epochs, enabling the pruning module to adjust domain partitions based on the evolving latent embeddings.
Our main contributions are:

\begin{itemize}[leftmargin=1.5em]
    \item To the best of our knowledge, this is the first work using medical ontologies to tackle clinical DG problems.
    It reveals the potential of medical ontologies for discovering latent domains for handling covariates, rather than serving as feature enrichment.
    \item \model{} shows robustness across four predictive tasks on two public datasets, outperforming both clinical DG~\citep{yang2023manydg, wu2023iterative} and regular DG baselines. For example, our model achieves up to a 12.8\% relative AUPRC improvement over the strongest baseline.
    \item We extend more analyses to show that \model{} handles domain shifts through hierarchy-grounded domain partitions and accurate invariant feature learning, without sacrificing computational overhead.
\end{itemize}

\section{Related Work}

\subsection{Domain Generalization (DG)}
DG is pursuing adjusted models which are specially designed to remove domain-covariate features from hidden representation~\citep{muandet2013domain}.
Most studies are dedicated to solve performance drops on target domain across diverse scenarios like image classification~\citep{zhou2022domain, ding2022domain, zhu2025pathology}, and they can be categorized as three different ways:
(i) An intuitive way is to minimize the empirical source risk, either domain alignment~\citep{ganin2016domain, li2018deep, zhao2020domain} and invariant learning technique~\citep{liu2021domain, zhang2022towards, wang2022variational} aim to convey little domain characteristics to acquire task-specific features.
(ii) Contrastive learning~\citep{kim2021selfreg, jeon2021feature, yao2022pcl} becomes an alternative for data augmentation, studies employed the contrastive loss function to reduce the gap of representation distribution in one category.
(iii) Meta-learning~\citep{balaji2018metareg, dou2019domain} and ensemble-learning~\citep{cha2021swad, chu2022dna} approaches handle domain shifts through dynamic loss functions.
However, they predefine either numerous or few domains in clinical settings~\citep{wu2023iterative}, which motivates us to design an adaptive way to discover latent domains from learnable parameters.

\subsubsection{DG in Clinical Prediction}
Prior works have shown that EHR predictive models often suffer performance drops when transferred to unseen patients with different data distributions~\citep{perone2019unsupervised, koh2021wilds}.
Some studies~\citep{zhao2020domain, zhang2021adaptive, hu2026exploring} adopt feature adaptation to align domain shifts across multiple hospitals and different time periods~\citep{reps2022learning, zhang2022adadiag}.
Recently, more works consider model generalization by mitigating the patient-specific domain shifts~\citep{guo2022evaluation, hai2024domain, zhu2024advancing}, which offers a flexible alternative in more scenarios.
Moreover, ManyDG~\cite{yang2023manydg} learns invariant features by treating each patient as unique domain; SLDG~\cite{wu2023iterative} develops a mixture-of-domain method to divide patient domains by concept-level features.
However, they address domain partition with simplifying heuristics such as linear dependencies~\citep{li2020domain}, while support from medical knowledge beyond training data remains unexplored in clinical DG problems.

\subsection{Hierarchy-Aware Predictive Modeling}
EHR prediction models can be broadly categorized into three types: RNN/CNN-based models~\citep{ChoiBSSS17, tong2025renaissance}, graph-based models~\citep{LuRCKN21, jiang2024graphcare}, and Transformer-based models~\citep{ShangMXS19,poulain2024graph, tong2025does, han2025black}.
Many graph-based studies~\citep{ChoiBSSS17, ShangMXS19, LuRCKN21} use medical hierarchies as structural priors, based on the assumption that diseases within the same cluster share common characteristics.
In practice, medical ontologies are often combined with co-occurrence graphs to define the message-passing structure of graph neural networks~\citep{xu2023seqcare, xu2024ram, hu2026bridging}.
However, this paradigm may introduce systematic bias, especially when disease associations capture local co-occurrence patterns rather than global clinical regularities~\citep{lin2026medcausalx}.
This limitation motivates us to study medical hierarchies beyond their common use for adjusting code embeddings.
Specifically, we explore the potential of medical hierarchies from the perspective of DG.
Hierarchical relations may help reveal how clinical patterns and data distributions vary across patient groups, yet such structure has rarely been used to characterize patient domains.
Our work aims to bridge this gap by integrating hierarchy-aware modeling with DG methods to improve the robustness of EHR predictive models under distribution shifts.

\begin{figure*}[t!]
    \centering
    \includegraphics[width=\linewidth]{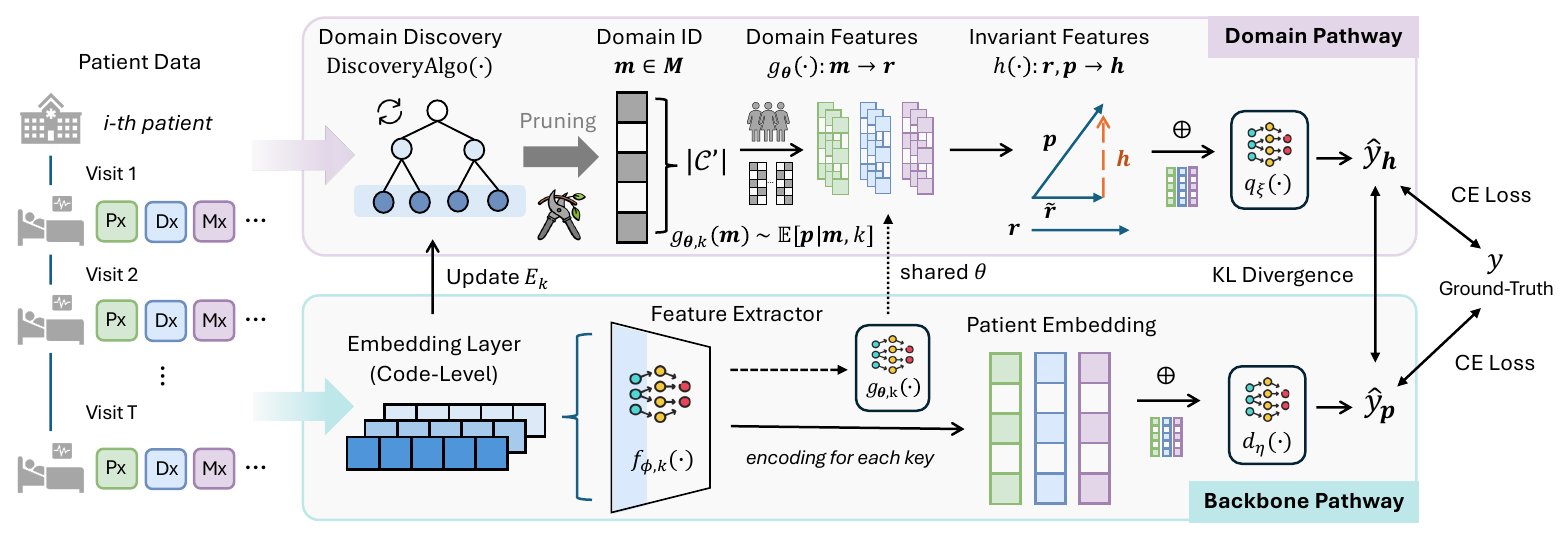}
    \caption{\textbf{The Overall Framework of \model{}.}
    The forward structure adds a domain pathway for mutual learning, extending beyond the backbone pathway of conventional predictive models. We feed patient data $\mathbf{x}$ into the backbone pathway, which learns patient features $\mathbf{p}$ through $f_{\phi,k}(\cdot)$ and produces the output prediction $\hat{y}_p$. In parallel, we obtain $\hat{y}_h$ along the domain pathway through $\mathrm{DiscoveryAlgo}(\cdot)$, $g_{\theta,k}(\cdot)$, and $h(\cdot)$. Here we can iteratively update domain partitions in $\mathbf{M}$ and update parameters on both pathways during training.
    }
    \Description{Framework diagram showing a backbone prediction pathway and a domain pathway that exchange information through iterative domain discovery and mutual learning.}
    \label{fig:framework}
\end{figure*}

\section{Preliminary}

\textbf{Clinical Prediction on EHR Data. \;}
Each patient $\mathbf{x^{(i)}}$ consists of a longitudinal sequence of admissions $\{V^{(i)}_1, V^{(i)}_2,\dots,V^{(i)}_T\}$.
We denote the entire set of medical concepts as \(\mathcal{C}=\{c_1, c_2,..., c_{|\mathcal{C}|}\}\), and each admission contains a subset of $\mathcal{C}$.
Most predictive models predict clinical outcomes $\mathbf{y}^{(i)} \in \{0,1\}^d$ at a future visit $V_{t+1}$, and they train a feature extractor $f_{\phi,k}(\cdot): \mathbf{x}_{k} \mapsto \mathbf{p}_{k}$ to encode $\mathbf{x}_{k}$ (omit $i$ in single patient) into patient embedding
\begin{align}
\mathbf{p}_{k} &= f_{\phi,k}(\mathbf{x}_k) \in \mathbb{R}^{h},
\label{eq:patient_embed}
\end{align}
where $f_{\phi,k}$ is the encoder for feature key $k$, and $h$ denotes the dimension.
When $f_{\phi}(\cdot)$ are trained on source training data sampled from distribution $P_{s}$, we often expect the learned model to also perform well on target data drawn from $P_t$ by assuming $P_s = P_t$.
However, it is possible that these two distributions differ a lot $P_s \neq P_t$ due to spatial and temporal shifts, which motivates the need for DG solutions to factorize label information that remains invariant across clinical environments.

\textbf{Concept-Specific Hierarchy. \;}
In EHR data, certain medical concept $c_i \in \mathcal{C}$ always originates from a hierarchical encoding system, such as International Classification of Diseases (ICD~\citep{world1988international}) and Anatomical Therapeutic Chemical (ATC~\citep{nahler2009anatomical}).
We define a concept-specific hierarchy $\mathcal{H}$ of $H$ levels, and $n^{(h)}_i$ is denoted as the $i$-th node on level $h$.
Leaf nodes at level $H$ represent actual codes via the mapping $m: c_i \mapsto n^{(H)}_i$ with node feature $\mathbf{e}_i$ stored in $f_{\phi,k}$, and $\mathrm{Desc}(\cdot)$ denotes the set of descendant nodes.
Note that, the root node $n^{(1)}_1$ at the top level subsumes all nodes in $\mathcal{H}$, and any two leaf nodes $n^{(H)}_i$ and $n^{(H)}_j$ share at least one common ancestor.
In this paper, we focus on the disease hierarchy (i.e. ICD-9-CM) for method clarity, while more ontologies (treatments and medications) are also involved and discussed in Appendix~\ref{app:with_more_features}.

\section{Methodology}
\model{} iteratively operates two main steps (as shown in Figure~\ref{fig:framework}): (1) developing a pruning algorithm for medical hierarchies to discover latent domains; (2) learning invariant features by factorizing patient embeddings in projection space.

\subsection{Hierarchy-Guided Domain Discovery}

While domain IDs are unavailable in EHR data, it is intuitive that patients with similar medical diagnoses (concepts) often belong to the same domain.
Naively partitioning patients by medical histories $\mathbf{x}$ implicitly induces soft domain labels, but leads to an exponential growth of latent domains with vocabulary size $|\mathcal{C}|$ (i.e., $2^{|\mathcal{C}|}$).
To address this issue, we propose a hierarchy-guided domain discovery algorithm that prunes overly fine-grained nodes and assigns domain IDs at appropriate ancestor levels, forming a compact domain set that still covers all concepts (see Figure~\ref{fig:intro}).
An assignment matrix $\mathbf{M}$ can be generated to query domain IDs via
\begin{equation}
\mathbf{M}
:=
\mathrm{DiscoveryAlgo}\Bigl(\{\mathbf{x}^{(i)}\}_{i=1}^{N_{\mathrm{tr}}}\Bigr)
\in
\{0,1\}^{N_{\mathrm{tr}}\times |\mathcal{C}^{'}|},
\label{eq:discovery}
\end{equation}
where \(N_{\mathrm{tr}}\) is the number of training patients, $|\mathcal{C^{'}}|$ is the pruned vocabulary size.
It merges fine-grained codes into higher-level clusters, allowing patients with or without a particular disease to occupy different latent domains as needed.

\subsubsection{Initialization on Domain IDs}
Following the previous settings~\citep{LuRCKN21, jiang2024graphcare}, we only focus on patients with multiple admissions $T \geq 2$.
This sequence should be then converted into a single vector that consolidates all prior visits.
Hence, we aggregate patient data $\{\mathbf{x}_1, \mathbf{x}_2, \dots, \mathbf{x}_T\}$ into a unified representation $\mathbf{X} = \bigvee_{t=1}^{T} \mathbf{x}_t$, where medical concepts shown in each admission are merged to form a accumulated medical history.
Note that, $\mathbf{X}$ is the most fine-grained domain partition for all patients.

\subsubsection{Initialization on Node Features}
With $\mathbf{X}$ for each patient, concept-specific ontology $\mathcal{H}$ with node features are required to decide whether fine-grained medical concepts group diseases into higher-level clusters or are preserved.
We initialize leaf-node features $\mathbf{e}_{i} \in \mathbb{R}^h$ by:
(1) for \textbf{present code} $c_i$ in $\mathcal{S}$, $\mathbf{e}_{i}$ is initialized from embedding layer $\text{E}(e_{1},\dots,e_{|\mathcal{C}|})$ in $f_{\phi,k}(\cdot)$;
(2) for \textbf{absent code} $c_i$ in $\mathcal{S}$, $\mathbf{e}_{i}$ is its embedding of entity name through ClinicalBERT~\citep{huang2019clinicalbert}.
The feature of a parent node $\mathbf{e}^{(h-1)}_{n_i}$ is then computed from the embeddings of its descendants at level $h$:
\begin{equation}
\mathbf{e}^{(h-1)}_i \;:=\; \frac{1}{|\mathrm{Desc}(n^{(h-1)}_{i})|} \sum_{n \in \mathrm{Desc}(n_{i})} \mathbf{e}^{(h)}_{n},
\label{eq:node_init}
\end{equation}
which extends $\text{E}(e_{1},\dots,e_{|\mathcal{C}|})$ to $\text{E}(e_1,\dots,e_{|\mathcal{H}|})$ over the entire hierarchy.
Still, it fails to capture the hierarchical distances in $\mathcal{H}$.
For example, two leaf nodes might have similar features without sharing the same parent node.
Under this scenario, their lowest common ancestor (LCA) $n_{\text{LCA}}(d_i,d_j)$ could be an alternative reflecting such similarity.
Hence, we mimic the principle of hierarchical clustering~\citep{johnson1967hierarchical} in this stage.
Concretely, we determine the most similar node pair $(d_i,d_j)$ based on their cosine similarity $\text{cos}(\mathbf{e}_{d_i}, \mathbf{e}_{d_j})$, and then update their LCA embedding
\begin{equation}
    \mathbf{e}_{\mathrm{LCA}(d_i,d_j)} \leftarrow \mathrm{Average}\bigl(\mathbf{e}_{\mathrm{LCA}(d_i,d_j)}, \mathbf{e}_{d_i}, \mathbf{e}_{d_j}\bigr).
\end{equation}
This process continues down the similarity list until the remaining highest similarity falls below the threshold $\rho=0.3$.

\label{sec:hierarchy-pruning}

\subsubsection{Node Scoring}
For each node $n\in \mathcal{H}$, a score function $S(n)$ is crucial for the algorithm to identify which node is a good ``candidate'' for final selection.
Motivated by information gains in Decision Tree~\citep{song2015decision}, we define three indicators for evaluation:
\begin{enumerate}
  \item \textbf{Coverage}, $\mathrm{cov}(n)$: measures the ratio of covered leaf nodes $n \in \mathcal{L}$.
  A higher value indicates greater coverage, favoring the selection of higher-level nodes.
  \item \textbf{Purity}, $\mathrm{pur}(n)$: gauges how cohesive the partitioned tree is in the embedding space.
  A higher value indicates that leaf nodes are similar to the subtree's root node.
  \item \textbf{Depth}, \(\mathrm{dep}(n)\in\{1,\dots,H\}\): indicating the level of node \(n\) in the hierarchy. A larger value corresponds to more fine-grained codes, favoring the selection of deeper nodes.
\end{enumerate}

We then assign a single score $S(n)$ to each node $n$:
\begin{align}
&S(n) = \exp(\mathrm{pur}(n)) +
\alpha \cdot \bigl( \mathrm{cov}(n) \times \mathrm{dep}(n) \bigr)
\notag \\
&= \exp \Bigl(\mathbb{E}_{m\in\mathcal{M}} [\mathrm{sim}(\mathbf{e}_n,\mathbf{e}_m )] \Bigr)
+ \alpha \cdot (\frac{|\mathcal{M}|}{|\mathcal{L}|} \cdot \frac{h}{H}),
\label{eq:score}
\end{align}
where $\mathcal{M}$ is equivalent to $\text{Desc(n)}$; $\mathbb{E}(\cdot)$ denotes the mathematical expectation; $\alpha$ and $\exp(\cdot)$ act as scaling factors, which regularize the selection avoiding either too low or high level.
The score matrix $S(s_1,\dots,s_{|\mathcal{H}|})$ is obtained after scoring all nodes in the hierarchy.

\subsubsection{Hierarchy Pruning}
Once $S(n)$ is computed, we perform a bottom-up pass over \(\mathcal{H}\) to generate a candidate set of pruned nodes.
Let $p$ be the parent node with children $\{c_1,\dots,c_r\}$.
There are three possible situations upon comparing certain node's score $S(p)$ with its children scores $\{S(c_i)\}^r$:

\textbf{Case 1.} If $S(p) > \max(\{S(c_i)\}^r)$, we include parent node \(p\) and exclude its children.

\textbf{Case 2.} If $S(p) < \min(\{S(c_i)\}^r)$, we discard parent node \(p\) and select all children.

\textbf{Case 3.} Otherwise, we tentatively include \(p\) but mark it for further resolution in the next step.

Note that, if $p$ is discarded in the above cases, $\{c_1,\dots,c_r\}$ are promoted to become the children of $p$'s parent.
After the first iteration on the score matrix, a candidate subset $\mathcal{C}_0$ is obtained; however, it remains sub-optimal and requires further evaluation.
To this end, we construct a list of tuples $A$ of length $N$, where each element $A[n] := (p, {c_1,\dots,c_r})$ represents a flagged parent--child pair.

\subsubsection{Domain Searching}
Given $N$ flagged pairs, the complexity of searching approaches grows exponentially $O(2^N)$.
Inspired by the Beam-Search algorithm~\citep{sutskever2014sequence}, we get decisions within the top-k pruning candidates $O(kN)$.
For each flagged pair, we either (1) unify them by including the parent or (2) retain the children as distinct pruned nodes.
Those pairs that have not yet been decided default to retaining parent node to ensure complete coverage.
We adopt the Silhouette Score~\citep{shahapure2020cluster} as the evaluation metric.
Specifically, \model{} first applies UMAP~\citep{mcinnes2018umap} on leaf node features for dimensionality reduction, and it then calculates the score on this full hierarchical partition, reflecting the global separation quality under the current pruning scheme.
This process iterates over all flagged node pairs in order by updating pruning subsets in $\{\mathcal{C}_1,\mathcal{C}_2,\dots, \mathcal{C}_N\}$ to make selection with the optimal evaluation result.

\subsubsection{Domain Decision}
With the resulting pruned selection $\mathcal{C}_N$ ($\mathcal{C}'$ in Eq.~\eqref{eq:discovery}), we update $\mathbf{M}$ by converting patient data into these pruned clusters in $\mathcal{C}_N$.
If patient $\mathbf{x}^{(i)}$ has at least one leaf code $d$ that descends from the pruned node, we update $\mathbf{M}[i,j] = 1$; otherwise, $\mathbf{M}[i,j] = 0$.
Consequently, each row $\mathbf{M}[i,:]$ represents the domain labels of the $i$-th patient, defining a structured domain partition over patients for downstream prediction.

\subsection{Mutual Forward Learning}
\label{sec:mutual-learning}

Each domain can be viewed as the latent representation $\mathbf{r}$ sampled from a meta domain distribution $p(\cdot)$, so that we can identify $\mathbf{r}$ and then factorize $p(y|\mathbf{x})$ into $\int p(y | \mathbf{x}, \mathbf{r}) p(\mathbf{r} | \mathbf{x}) \, d\mathbf{r}$ by approximating $q(\mathbf{r}) \sim p(\mathbf{r}|\mathbf{x})$ given data samples $\mathbf{x}$.
Subsequently, a domain encoder $p(\mathbf{r}|\mathbf{x})$ and a label predictor $p(y|\mathbf{x},\mathbf{r})$ are needed for inference.
Here we parameterize the domain encoder $p(\mathbf{r}|\mathbf{x})$ as a network $g_\theta(\cdot)$ with parameter $\theta$.
Since the pruned output matrix $\mathbf{M}$ (see Section~\ref{sec:hierarchy-pruning}) maps each training sample $\mathbf{x}$ to $\mathbf{m}$ (soft-label domain IDs), we apply $g_\theta(\cdot)$ to $\mathbf{m}$ to estimate the domain factor $\mathbf{r} := g_\theta(\mathbf{m})$.
Although $\mathbf{r}$ represents a probabilistic domain variable, we implement $g_\theta$ as a deterministic Multi-Layer Perceptron (MLP) for the prediction task.
Next, we compute invariant features using a non-parametric function $h(\cdot): (\mathbf{r}, \mathbf{p}) \mapsto \mathbf{h}$, which disentangles domain covariates $\mathbf{r}$ from patient embeddings $\mathbf{p}$ 
to obtain features for predicting $p(y|\mathbf{x}, \mathbf{r})$.

\subsubsection{Self-Supervised Domain Encoder}
The main concerns on training domain encoder is how to ensure $g_\theta(\cdot)$ can extract valid domain information from patients, which is ignored by some works~\citep{finn2017model, li2018learning, yang2023manydg}.
A regulation method is then developed during the encoder training phase.
Concretely, pseudo domain labels $\mathbf{m}$ help us divide patients into latent domains, where averaging patient-specific features $\bar{\mathbf{p}}$ could provide guidance for $g_\theta(\cdot)$ in identifying domain information.
Hence, we adopt a pretraining task and update $\theta$ based on patient embeddings $\mathbf{p}$ from $f_\phi(\cdot)$ by minimizing loss function
\begin{align}
\label{eq:loss_domain_encoder}
    \mathcal{L}_r [g_\theta(\mathbf{m}),\bar{\mathbf{p}}]
    := \mathrm{MSE}(\mathbf{r},\;\mathbb{E}[\mathbf{p} | \mathbf{m}]) + \frac{\|\mathbf{r}_{\mu} - \mathbf{p}_{\mu} \|_{\mathcal{F}}^2}
     {\|\mathbf{p}_{\mu} \|_{\mathcal{F}}^2},
\end{align}
where $\mathbb{E}[\mathbf{p} | \mathbf{m}]$ denotes the average embedding associated with domain IDs, and $\|\mathbf{r}_{\mu} - \mathbf{p}_{\mu}\|_{\mathcal{F}}^2$ measures the Maximum Mean Discrepancy~\citep{borgwardt2006integrating} with the norm $\mathcal{F}$ to reduce distributional gaps.
The subscript $\mu$ indicates batch-level averages.
The domain encoder can then approximate $\mathbf{r}$ through both patient-specific inputs $\mathbf{p}$ and $\mathbf{m}$.

\subsubsection{Invariant Feature Projection Learning}
In Eq.~\eqref{eq:loss_domain_encoder}, both $\mathbf{r}$ and $\mathbf{p}$ are rescaled into a shared vector space with comparable magnitudes.
Hence, we can directly apply an orthogonal projection approach (as in early studies~\citep{bousmalis2016domain,shen2022connect,yang2023manydg}) to obtain the invariant feature $\mathbf{h}$ by subtracting the parallel component of $\mathbf{p}$ in this shared vector space.
We formalize this in $h(\cdot)$ as shown in Eq.~\eqref{eq:vector_decomposition}:
\begin{align}
    \label{eq:vector_decomposition}
    \mathbf{h} := \mathbf{p} - \tilde{\mathbf{r}}, \;\text{where} \; \tilde{\mathbf{r}} = \mathbf{r} \cdot \langle \frac{\mathbf{p}}{\|\mathbf{r}\|}, \frac{\mathbf{r}}{\|\mathbf{r}\|} \rangle.
\end{align}
Here, $\tilde{\mathbf{r}}$ is the component of $\mathbf{p}$ that is parallel to $\mathbf{r}$ with domain covariates, while $\mathbf{h}$ is the remainder and thus invariant to domain shifts.
We thus obtain $\mathbf{h}$ without additional parameters, and $h(\cdot)$ serves as label information before making prediction.

\begin{algorithm}[t!]
  \caption{Overview of \model{}}
  \label{alg:udoncare}
  \begin{algorithmic}[1]
    \STATE {\bfseries Input:} EHR dataset $\mathcal{S}$ with patient's data $\{\mathbf{x}^{(i)}\}_{i=1}^{N_{\mathrm{tr}}}$; Feature extractor $f_{\phi}$ with defined backbone (e.g. Transformer).

    \STATE \textcolor{cyan}{// \textit{When iteration $I=3$ and epochs $N=100$}}
    \FOR{epoch $\in \{1,2,\dots,40\}$}
      \STATE \textcolor{cyan}{// \textit{Backbone Pathway}}
      \STATE Decode $\hat{y}_p \gets d_\eta(f_{\phi,k}(x))$ with Equation~\eqref{eq:label_predictor};
    \ENDFOR

    \STATE Obtain learned $\phi$; Initialize the hierarchy $\mathcal{H}$;

    \FOR{iteration $\in \{1,2,3\}$}
      \STATE \textcolor{cyan}{// \textit{Hierarchy-Guided Domain Discovery}}
      \STATE Define \& Update look-up table $\mathbf{M}$ via Step 1-4;
      \STATE Assign domain IDs $\mathcal{M}: \mathbf{x} \mapsto \mathbf{m}$ with Equation~\eqref{eq:discovery};

      \STATE \textcolor{cyan}{// \textit{Self-Supervised Domain Encoding}}
      \STATE Initialize $g_{\theta}$; Pretrain $g_{\theta}$ by minimizing Equation~\eqref{eq:loss_domain_encoder};

      \FOR{epoch $\in \{1,2,\dots,20\}$}
        \STATE \textcolor{cyan}{// \textit{Domain Pathway}}
        \STATE Obtain patient embeddings $\mathbf{p}$ with Equation~\eqref{eq:patient_embed};
        \STATE Get domain features $\mathbf{r}$ by $g_{\theta}(\mathbf{m})$;
        \STATE Decompose invariant features $\mathbf{h}$ with Equation~\eqref{eq:vector_decomposition};
        \STATE Decode $\hat{y}_p, \hat{y}_h$ with Equation~\eqref{eq:label_predictor};
        \STATE Minimize co-training loss $\mathcal{L}_p, \mathcal{L}_h$ with Equation~\eqref{eq:co-training};
      \ENDFOR
    \ENDFOR

    \STATE {\bfseries Output:} Trained models $f_{\phi}$, $g_{\theta}$, $d_{\eta}$, $q_{\xi}$; final prediction $\hat{y}_h$.
  \end{algorithmic}
\end{algorithm}

\subsection{Training and Inference}
\label{sec:train-inference}

\subsubsection{Iterative Training.}
To train \model{}, we feed each data sample $\mathbf{x}$ into the hierarchy-pruning module to obtain its domain ID $\mathbf{m}$, and then remove domain covariates under a mutual learning architecture.
Rather than updating the model continuously in each epoch, we adopt an iterative training strategy, which prior studies~\citep{cui2019deep, sofiiuk2022reviving} have shown can reduce training time while maintaining comparable predictive performance.\footnote{For example, we set iterations $I = 3$ and epochs $N=100$ by obtaining pretrained parameters for 40 epochs and then updating $\mathbf{M}$ iteratively every 20 epochs, yielding a total of 100 epochs.}
We iteratively update the model weights and regenerate domain assignments every 20 epochs in our experiment.
Before each iteration, we reinitialize $g_\theta(\cdot)$, because the input shape of $\mathbf{m}$ may vary due to updated code-level embeddings.
The full procedure is summarized in Algorithm~\ref{alg:udoncare}.

\subsubsection{Mutual Inference.}
After the orthogonal projection, we apply the network $q_\xi(\cdot)$ (operates on space $\mathbf{p}$) as a post-step to parameterize the label predictor $p(y|\mathbf{x}, \mathbf{r})$;
We can also parameterize $p(y|\mathbf{x})$ through the regular decoder network $d_\eta(\cdot)$ fed by patient embeddings from backbone $f_\phi(\cdot)$.
\begin{align}
\label{eq:label_predictor}
    p(y|\mathbf{x}) \sim \hat{y}_p &=
    d_\eta(\mathbf{p}) =
    d_\eta(f_\phi(\mathbf{x})),
    \notag \\
    p(y|\mathbf{x}, \mathbf{r}) \sim\hat{y}_h &=
    q_\xi(\mathbf{h}) =
    q_\xi(h(g_\theta(\mathbf{m}), \mathbf{p})).
\end{align}
Both $q_\xi(\cdot)$ and $d_\eta(\cdot)$ are MLP classifiers with learnable parameters.
A loss function $\mathcal{L}$ is then applied to add label supervision for downstream predictive tasks.
We integrate these two predictors into a collaborative framework, with the mutual inference objective:
\begin{align}
    \label{eq:co-training}
    \mathcal{L}_p &= \mathbb{E}_{(\mathbf{x}, y)\sim\mathcal{S}_{train}}\ell(\hat{y}_{p}, y)+\lambda \cdot D_{\text{KL}}(\hat{y}_{p} ||\; \tilde{y}),
    \notag \\
    \mathcal{L}_h &= \mathbb{E}_{(\mathbf{x}, y)\sim\mathcal{S}_{train}}\ell(\hat{y}_{h}, y)+\lambda
    \cdot D_{\text{KL}}(\hat{y}_{h} ||\; \tilde{y}),
\end{align}
where $D_{\text{KL}}(\hat{y}_{*} \,\|\, \tilde{y})$ denotes the KL Divergence~\citep{van2014renyi} with the same $\lambda$ value in $\mathcal{L}_p$ and $\mathcal{L}_h$. Within each loss, $\ell(\cdot)$ denotes the binary cross entropy and $\tilde{y}$ is the average probability of $\hat{y}_p$ and $\hat{y}_h$. These two losses are calculated jointly $\mathcal{L}_p+\mathcal{L}_h$ to let $d_\eta$ and $q_\xi$ regularize one another, stabilizing the learning of $q_\xi$.
Following the DG setting, \model{} applies $\hat{y}_h$ as the final prediction.

\section{Experiments}
\subsection{Experiment Setups}

\begin{table}[!t]
\centering
\caption{Statistics of MIMIC-III and MIMIC-IV datasets.}
\begin{tabular}{lcc}
\toprule
\textbf{Dataset} & \textbf{MIMIC-III} & \textbf{MIMIC-IV} \\
\midrule
\# patients & 6,497 & 49,558 \\
Max. \# visit & 42 & 55 \\
Avg. \# visit & 2.66 & 3.66 \\
\midrule
Avg. \# conditions per visit & 13.06 & 13.38 \\
Avg. \# procedures per visit & 4.54 & 4.70 \\
Avg. \# medications per visit & 33.71 & 43.89 \\
\bottomrule
\end{tabular}
\label{tab:basic_stat}
\end{table}

\begin{table*}[t]
\centering
\caption{\textbf{Performance comparison of four prediction tasks on MIMIC-III/MIMIC-IV.} We report the average performance (\%) and the standard deviation (in bracket) over 5 runs. We also \textbf{bold} for the best model.}
\label{tab:main_result}
\resizebox{0.92\textwidth}{!}{
\begin{tabular}{lcccccccc}
\toprule
\multirow{3}{*}{\textbf{Model}}
&\multicolumn{4}{c}{\textbf{Task 1: Mortality Prediction}} & \multicolumn{4}{c}{\textbf{Task 2: Readmission Prediction}} \\
& \multicolumn{2}{c}{\textbf{MIMIC-III}} & \multicolumn{2}{c}{\textbf{MIMIC-IV}} & \multicolumn{2}{c}{\textbf{MIMIC-III}} & \multicolumn{2}{c}{\textbf{MIMIC-IV}} \\
\cmidrule(lr){2-3} \cmidrule(lr){4-5} \cmidrule(lr){6-7} \cmidrule(lr){8-9}
& \textbf{AUPRC} & \textbf{AUROC} & \textbf{AUPRC} & \textbf{AUROC} & \textbf{AUPRC} & \textbf{AUROC} & \textbf{AUPRC} & \textbf{AUROC} \\
\midrule
Oracle & 16.95\;(0.48) & 70.61\;(0.58) & 8.46\;(0.53) & 68.92\;(0.47) & 74.42\;(0.43) & 69.74\;(0.48) & 69.37\;(0.12) & 68.47\;(0.11) \\
\midrule
Base & 11.05\;(0.65) & 55.01\;(0.98) & 3.97\;(0.47) & 59.13\;(0.76) & 63.45\;(0.84) & 62.15\;(0.73) & 58.46\;(0.24) & 60.56\;(0.31) \\
DANN & 12.80\;(0.52) & 63.29\;(0.69) & 4.41\;(0.45) & 63.82\;(0.46) & 64.02\;(0.61) & 63.89\;(0.75) & 60.89\;(0.11) & 61.22\;(0.23) \\
CondAdv & 13.60\;(0.52) & 65.22\;(0.53) & 5.65\;(0.62) & 64.27\;(0.68) & 67.72\;(0.41) & 65.47\;(0.50) & 62.78\;(0.14) & 62.29\;(0.19) \\
MLDG & 13.25\;(0.44) & 64.07\;(0.63) & 4.75\;(0.38) & 62.75\;(0.57) & 66.00\;(0.55) & 65.59\;(0.59) & 60.20\;(0.23) & 62.24\;(0.28) \\
IRM & 13.50\;(0.48) & 65.41\;(0.54) & 4.14\;(0.52) & 62.36\;(0.61) & 67.07\;(0.43) & 65.93\;(0.55) & 61.81\;(0.14) & 61.84\;(0.17) \\
PCL & 13.70\;(0.49) & 64.55\;(0.61) & 5.35\;(0.49) & 64.70\;(0.55) & 65.03\;(0.49) & 66.36\;(0.59) & 61.03\;(0.28) & 63.04\;(0.26) \\
ManyDG & 14.10\;(0.54) & 66.27\;(0.52) & 6.06\;(0.31) & 64.66\;(0.32) &68.63\;(0.45) & 67.03\;(0.56) & 63.48\;(0.25) & 64.07\;(0.24) \\
SLDG   & 13.20\;(0.47) & 63.63\;(0.56) & 4.58\;(0.44) & 63.24\;(0.60) & 65.73\;(0.53) & 65.80\;(0.50) & 61.55\;(0.14) & 65.29\;(0.16) \\
\midrule
\model{} & \textbf{15.90\;(0.36)} & \textbf{68.78\;(0.40)} & \textbf{6.81\;(0.27)} & \textbf{66.73\;(0.48)} & \textbf{71.17\;(0.32)} & \textbf{67.68\;(0.36)} & \textbf{65.61\;(0.10)} & \textbf{67.26\;(0.25)} \\
\midrule
\multirow{3}{*}{\textbf{Model}}
&\multicolumn{4}{c}{\textbf{Task 3: Drug Recommendation}} & \multicolumn{4}{c}{\textbf{Task 4: Diagnosis Prediction}} \\
& \multicolumn{2}{c}{\textbf{MIMIC-III}} & \multicolumn{2}{c}{\textbf{MIMIC-IV}} & \multicolumn{2}{c}{\textbf{MIMIC-III}} & \multicolumn{2}{c}{\textbf{MIMIC-IV}} \\
\cmidrule(lr){2-3} \cmidrule(lr){4-5} \cmidrule(lr){6-7} \cmidrule(lr){8-9}
& \textbf{AUPRC} & \textbf{F1-score} & \textbf{AUPRC} & \textbf{F1-score} & \textbf{w-F$_1$} & \textbf{R@10} & \textbf{w-F$_1$} & \textbf{R@10} \\
\midrule
Oracle & 80.43\;(0.14) & 67.41\;(0.29) & 74.31\;(0.25) & 61.28\;(0.22) & 26.97\;(0.13) & 39.46\;(0.27) & 28.12\;(0.11) & 40.53\;(0.16) \\
\midrule
Base & 68.83\;(0.19) & 57.31\;(0.35) & 66.94\;(0.18) & 53.13\;(0.18) & 21.32\;(0.19) & 31.08\;(0.23) & 20.07\;(0.12) & 31.52\;(0.18) \\
DANN & 75.18\;(0.15) & 60.55\;(0.30) & 69.63\;(0.27) & 53.43\;(0.26) & 22.08\;(0.15) & 34.27\;(0.28) & 24.05\;(0.14) & 35.21\;(0.20) \\
CondAdv & 76.59\;(0.18) & 64.42\;(0.25) & 71.48\;(0.15) & 55.62\;(0.29) & 23.09\;(0.18) & 36.27\;(0.21) & 26.13\;(0.12) & 37.35\;(0.18) \\
MLDG & 74.76\;(0.17) & 58.85\;(0.33) & 70.29\;(0.27) & 56.77\;(0.16) & 21.34\;(0.17) & 34.16\;(0.26) & 24.17\;(0.13) & 34.72\;(0.19) \\
IRM & 69.47\;(0.16) & 62.20\;(0.30) & 69.12\;(0.14) & 54.57\;(0.18) & 22.27\;(0.16) & 32.86\;(0.23) & 23.54\;(0.15) & 34.12\;(0.21) \\
ManyDG & 77.29\;(0.14) & 63.70\;(0.28) & 71.26\;(0.19) & 55.27\;(0.19) &23.39\;(0.14) & 36.49\;(0.27) & 25.91\;(0.13) & 37.04\;(0.20) \\
\midrule
\model{} & \textbf{78.16\;(0.18)} & \textbf{66.70\;(0.29)} & \textbf{73.07\;(0.33)} & \textbf{59.23\;(0.14)} & \textbf{24.11\;(0.18)} & \textbf{38.37\;(0.26)} & \textbf{27.31\;(0.11)} & \textbf{39.41\;(0.17)} \\
\bottomrule
\end{tabular}
}
\end{table*}

\subsubsection{Predictive Tasks}
We evaluate \model{} on four representative tasks:
(1) \textbf{Mortality Prediction}, which determines whether a patient will pass away by a specified time horizon after discharge. This is a binary classification task.
(2) \textbf{Readmission Prediction}, which checks if a patient will be readmitted within a predefined window (e.g., next 15 days) following discharge. This is also framed as a binary classification.
(3) \textbf{Diagnosis Prediction}, which forecasts the set of diagnoses (ICD-9-CM codes) for the patient's next hospital visit based on prior visits. This requires multi-label classification.
(4) \textbf{Drug Recommendation}, which suggests a set of medications (ATC-4 codes~\citep{nahler2009anatomical}) for the upcoming visit, also formulated as multi-label classification.
These tasks reflect diverse clinical needs and provide a rigorous benchmark for DG evaluation.

\subsubsection{Evaluation Metrics}
Both readmission and mortality prediction are binary classification tasks, we calculate the Area Under the Precision-Recall Curve (AUPRC) and the Area Under the Receiver Operating Characteristic Curve (AUROC) scores due to the imbalanced label distribution.
For drug recommendation, we evaluate predictions of all DG approaches by AUPRC and F1 scores, the same setting as ManyDG~\citep{yang2023manydg} (i.e. $d < 120$).
For diagnosis prediction, we decide accurate prediction by weighted $\text{F}_1$ score as in Timeline and top-10 recall as in DoctorAI~\citep{choi2016doctor}, since the former one measures the overall prediction on all classes (i.e. $d > 4500$) and the latter one have concentration on positive code with low frequency.

\subsubsection{Datasets \& Data Split}
We conduct our main experiments on \textbf{MIMIC-III} and \textbf{MIMIC-IV}, which are widely used in clinical prediction~\citep{johnson2016mimic, johnson2023mimic}.
MIMIC-III covers ICU admissions from 2001 to 2012, while MIMIC-IV spans 2008 to 2019.
To avoid overlapping time ranges with MIMIC-III, we only retain patients from the years $2013$--$2019$ in MIMIC-IV.
For each set of experiments, we extract $6{,}497$ and $49{,}558$ patients with multiple visits ($T \ge 2$) from both datasets as shown in Table~\ref{tab:basic_stat}.
Different from random data splitting, we evaluate our model under domain shifts induced by demographic and temporal factors. For the MIMIC-III dataset, patients with Medicare or Medicaid insurance are used as the source domain, while those with private insurance are treated as the target domain. Following the setting in SLDG~\citep{wu2023iterative} on the MIMIC-IV dataset, patients admitted after 2017 are assigned to the target domain, while all preceding patients are used as the source domain. 
Note that, there is no patient overlap between the source and target domains to prevent information leakage.

\subsubsection{Baselines}
We compare \model{} with two na\"ive baselines, five general DG baselines and two most recent clinical DG baselines:
(1) Na\"ive Baselines: \textbf{Oracle}, trained backbone encoder directly on the target data, and \textbf{Base}, trained solely on the source data.
The difference between metrics from Oracle (upper bound) and Base (lower bound) can show the distribution gaps between source and target data.
(2) Typical DG Baselines: \textbf{DANN}~\citep{ganin2016domain},
\textbf{CondAdv}~\citep{li2018deep},
\textbf{MLDG}~\citep{li2018learning},
\textbf{IRM}~\citep{arjovsky2019invariant}, and
\textbf{PCL}~\citep{yao2022pcl}.
(3) Clinical DG Baselines:
\textbf{ManyDG}~\citep{yang2023manydg}, and
\textbf{SLDG}~\citep{wu2023iterative}.
Since drug recommendation and diagnosis prediction are multi-label classification, we drop PCL and SLDG in these two tasks due to their setting limitation.
Note that, all baselines follow the same source/target definition as in the Data Split section, and the test set remains unseen to all models except Oracle.
The results reported in following sections are the performance evaluated on the test set.

\subsubsection{Implementation Details. }
Considering the common use of the Encoder-Decoder structure for clinical prediction, we adopt the Transformer~\citep{vaswani2023attentionneed} as the backbone feature extractor $f_\phi$ in \model{} and all baselines.
Specifically, we follow the implementation adapted from PyHealth~\citep{yang2023pyhealth}, consisting of three layers with a hidden size of 64, 4 attention heads, and a dropout rate of 0.2.
The position encoding is applied across patient visits to capture temporal order. Diagnosis, treatment, and medication codes are embedded as separate feature keys using an embedding look-up table.
The MLP classifiers used for both original and domain-invariant representations contain two hidden layers with sizes of 64 and 32, ReLU activations, and a dropout rate of 0.2, with task-specific output activations.
We use the Adam optimizer for training, and all remaining hyperparameters follow the settings in PyHealth.
For domain discovery, the score function uses $\alpha = 0.5$, and the KL divergence loss coefficient $\lambda$ is set to 1.0 on MIMIC-III and 1.5 on MIMIC-IV.
All models are trained for 100 epochs, and the best model is selected based on the AUPRC score monitored on the source validation set.
We set the learning rate to $1 \times 10^{-4}$ for $f_\phi$ and $5 \times 10^{-5}$ for $g_\theta$, batch size to 32, iteration to 3, and self-supervised epoch to 30.
All experiments are conducted using Python 3.10 and PyTorch 2.3.1 with CUDA 12.4 on a server equipped with AMD EPYC 9254 24-Core Processors and NVIDIA L40S GPUs.
Code is available at \url{https://github.com/humphreyhuu/UdonCare}.

\begin{figure}[t]
    \centering
    \includegraphics[width=\linewidth]{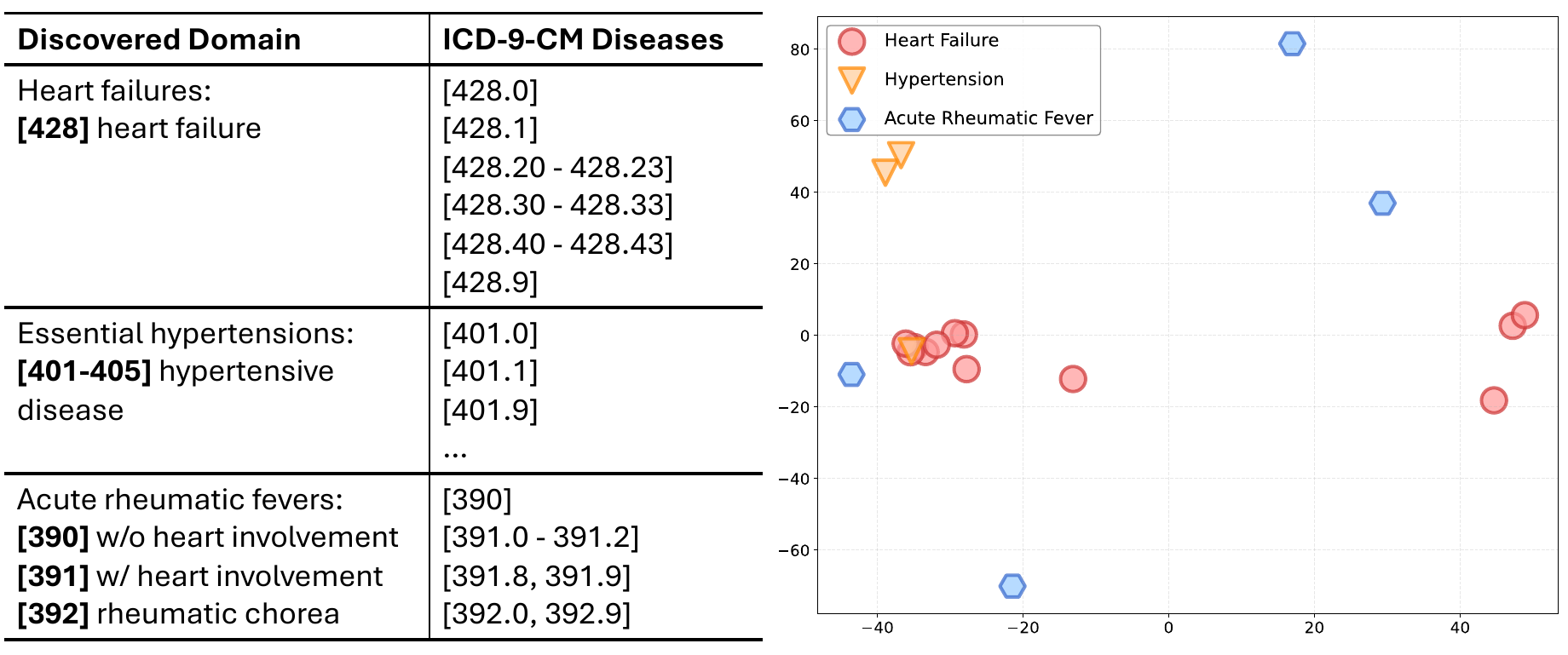}
    \caption{\textbf{Domains among diseases of the circulatory system identified by \model{}.} The left table shows how the discovered domains correspond to specific ICD-9-CM diseases (codes) in the hierarchy, while the right figure is the scatter plot using t-SNE upon code-level embeddings.}
    \Description{Table and t-SNE scatter plot showing discovered circulatory-system disease domains and their corresponding ICD-9-CM code embeddings.}
    \label{fig:interpret}
\end{figure}

\subsection{Main Results}
\label{sec:main-result}
Table~\ref{tab:main_result} presents results on four classification tasks using MIMIC-III and MIMIC-IV. First, the performance gap between the Oracle and Base methods is substantial, showing the presence of considerable domain differences.
DANN~\citep{ganin2016domain} and MLDG~\citep{li2018learning}, both relying on coarse domain partitions, show minimal improvements, likely due to difficulties in extracting consistent features from coarse partitions.
PCL~\citep{yao2022pcl} exhibits a slight gain through proxy-to-sample relationships.
Meanwhile, IRM~\citep{arjovsky2019invariant} and CondAdv~\citep{li2018deep} perform better by incorporating regularization or recurrent structures for predictions.
Among clinical-specific baselines, ManyDG~\citep{yang2023manydg} achieves the best results by leveraging mutual reconstruction, and SLDG~\citep{wu2023iterative} sees only modest improvement due to its reliance on the most recent admissions.
Lastly, \model{} surpasses over all baselines across all tasks.
Specifically, \model{} achieves relative AUPRC improvements of 12.8\% (12.4\%) for mortality prediction and 3.7\% (3.4\%) for readmission prediction on MIMIC-III (MIMIC-IV), respectively.

Beyond the above results, we further extend our experiments by using GAMENet~\citep{shang2019gamenet} as our backbone, and evaluate model generalization across hospitals on the eICU dataset (see Appendix~\ref{app:supp_main}).

\subsection{Analysis of Discovered Domain}

To illustrate how \model{} behaves, we evaluate discovered domains on diagnosis prediction using MIMIC-III.
Here, we focus on three kinds of heart related conditions from diseases of the circulatory system: heart failure (HF, 15 types), essential hypertension (EH. 3 types), and acute rheumatic fever (ARF, 7 types).
In the left table of Figure~\ref{fig:interpret}, we can observe that the mid-level ``heart failure'', high-level ``hypertensive disease'', and several leaf-level nodes are chosen to represent these subtypes.
These decisions reflect the same patterns of code-embedding geometry (right scatter plot): HF codes form a compact cluster, EH codes cluster even more tightly and lie near the HF group, while ARF codes are widely dispersed.
Moreover, it is also consistent with medical knowledge: HF and EH are clinically cohesive, while ARF shows greater heterogeneity due to variable cardiac involvement.
Thus, these results align with both data-driven and clinical logic.

\begin{figure}[t]
    \centering
    \includegraphics[width=0.9\linewidth]{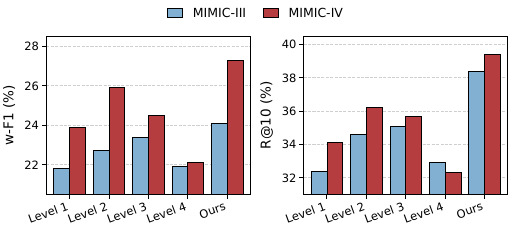}
    \caption{\textbf{Diagnosis-prediction comparison across five partitioning schemes.} Levels~1--4 use a globally fixed ICD-9-CM hierarchy depth, while \model{} adaptively prunes per branch.}
    \Description{Grouped bar charts of diagnosis-prediction w-F1 and R@10 under five partition schemes on MIMIC-III and MIMIC-IV.}
    \label{fig:partition-pred}
\end{figure}

\subsection{Effectiveness of Domain Partitions.}
\begin{figure}[t]
    \centering
    \includegraphics[width=0.8\linewidth]{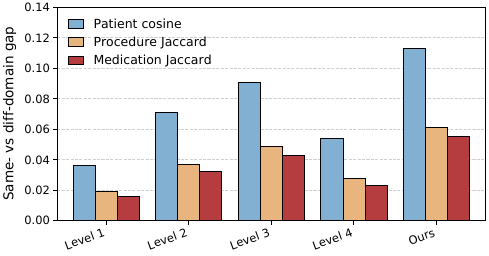}
    \caption{\textbf{Held-out patient coherence under five partitioning schemes.} For each scheme, we sample same-domain and different-domain held-out patient pairs and report their similarity gap on frozen patient embeddings, procedure codes, and medication codes.}
    \Description{Grouped bar chart of held-out patient coherence gaps across three modalities under five partition schemes. }
    \label{fig:partition-coh}
\end{figure}

A natural alternative to adaptive pruning is to commit to a single fixed depth of the ICD-9-CM hierarchy as the partition.
We compare \model{} against four such fixed-level baselines on diagnosis prediction, ranging from coarse chapters (Level~1, e.g., 390--459) to leaf codes (Level~4, e.g., 428.1).
Figure~\ref{fig:partition-pred} shows that fixed aggregation is a meaningful baseline: intermediate levels~2--3 outperform either extreme, but no single level wins across datasets.
\model{} consistently surpasses all fixed levels, indicating that the additional gain comes from per-branch granularity selection rather than from hierarchy usage alone.
A stronger partition should also define more coherent patient groups: following a balanced pairwise-coherence protocol, we sample same-domain and different-domain held-out patient pairs under each scheme and report the similarity gap on frozen patient embeddings, procedure codes, and medication codes (Figure~\ref{fig:partition-coh}).
\model{} yields the largest gap across all three modalities, suggesting that its discovered domains correspond to clinically coherent strata rather than arbitrary regrouping of diagnosis codes.

\begin{table*}[!t]
  \centering
  \caption{\textbf{Cosine similarity of linear weights on $\mathbf{p}$, $\widetilde{\mathbf{r}}$, and $\mathbf{h}$.} Cosine similarity is averaged across classes and evaluated over 3 runs.}
  \label{tab:cosine-similarity}
  \resizebox{\textwidth}{!}{
  \begin{tabular}{lcccc}
    \toprule
    Cosine Similarity & Mortality Prediction & Readmission Prediction & Drug Recommendation & Diagnosis Prediction \\
    \midrule
    $W_d(\mathbf{p}\rightarrow \text{labels})$ vs.\ $W_d(\mathbf{h}\rightarrow \text{labels})$ &
      $0.7831 \pm 0.0174$ & $0.4813 \pm 0.0391$ & $0.6546 \pm 0.0237$ & $0.8654 \pm 0.0198$ \\
    $W_d(\mathbf{p}\rightarrow \text{domains})$ vs.\ $W_d(\widetilde{\mathbf{r}}\rightarrow \text{domains})$ &
      $0.3427 \pm 0.0088$ & $0.1957 \pm 0.0314$ & $0.2753 \pm 0.0274$ & $0.2133 \pm 0.0036$ \\
    $W_d(\mathbf{p}\rightarrow \text{labels})$ vs.\ $W_d(\mathbf{p}\rightarrow \text{domains})$ &
      $0.1239 \pm 0.0126$ & $0.0794 \pm 0.0392$ & $0.1251 \pm 0.0212$ & $0.0972 \pm 0.0095$ \\
    \bottomrule
  \end{tabular}
  }
  \footnotesize
  $*$ $W_{\text{d}}(\cdot)$ represents the learned linear weights. For example, $W_{\text{d}}(\mathbf{p}\rightarrow \text{domains})$ denotes training a linear model on $\mathbf{p}$ to predict domains.
\end{table*}

\subsection{Ablation Analysis}
\begin{figure}[!t]
    \centering
    \includegraphics[width=0.9\linewidth]{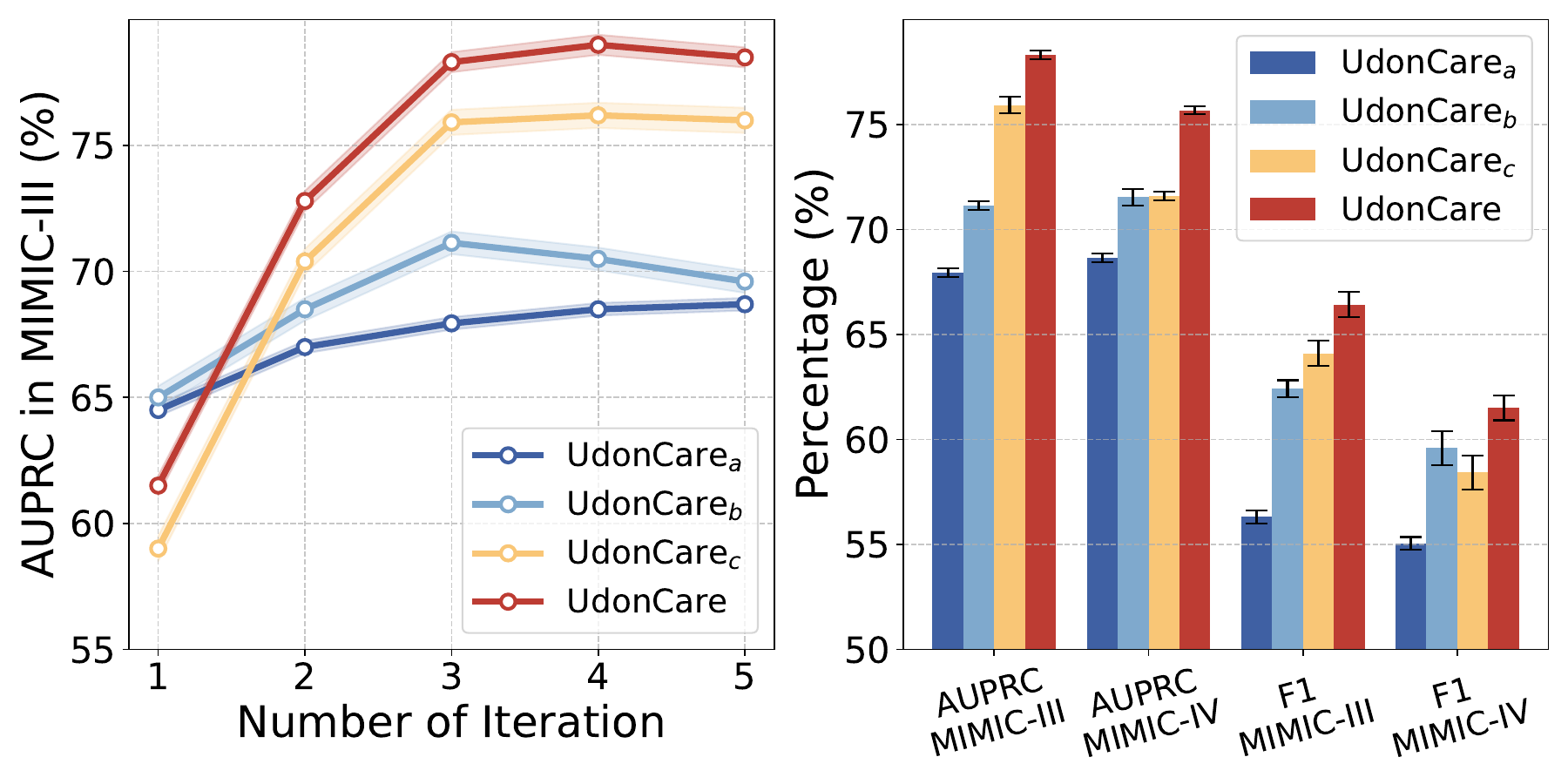}
    \caption{\textbf{Effectiveness of Domain Discovery.} The left figure shows the effect of the number of iteration on AUPRC, and the right one shows comparison among variants upon \model{}.}
    \Description{Two-panel plot comparing AUPRC across discovery iterations and model variants for domain discovery.}
    \label{fig:case_a}
\end{figure}

We evaluate whether the designed pruning algorithm is effective for prediction.
We adopt (a) k-Means clustering, (b) hierarchical clustering, and (c) tree pruning of the information gain algorithms as alternatives to simplify the partitioning process.
We use drug recommendation tasks as an example to assess performance.
The results can be found in Figure~\ref{fig:case_a}.

The left figure shows how AUPRC changes with the number of domain-discovery iterations.
We observe that k-Means performs the worst, largely because of its limitations in determining the optimal number of clusters via grid search.
Hierarchical clustering performs better than k-Means but lacks structured guidance from the medical hierarchy.
Tree pruning outperforms both methods by leveraging medical ontologies, demonstrating the importance of knowledge-driven clustering.
Moreover, \model{} outperforms all these methods by incorporating more precise domain IDs through iterative beam-search updates.
These results highlight the critical role of medical ontologies in domain discovery and the advantages of adaptive refinement for learning meaningful structures.

\subsection{Effectiveness of Decomposition}
\label{sec:probing}
Following the probing setting of~\cite{shen2022connect, yang2023manydg}, a linear classifier $d$ can be trained on the embedding to predict either (i) labels or (ii) domains. After training such a predictor, the cosine similarity can be calculated in terms of the learned weights to quantify the feature dimension overlaps.
Note that $\mathbf{p}$, $\widetilde{\mathbf{r}}$, and $\mathbf{h}$ are normalized dimension-wise to ensure that each dimension is comparable.
The results are shown in Table~\ref{tab:cosine-similarity}.
In general, the third row has lower cosine similarities than the first two rows, which indicates that there is mostly non-overlap between feature dimensions predicting labels and domains.
Moreover, the first two rows give relatively higher similarity and imply the domain and label information are separated from $\mathbf{p}$ into domain features $\tilde{\mathbf{r}}$ (scaling from $\mathbf{r}$) and invariant features $\mathbf{h}$.
It provides the quantitative evidence that \model{} stores domain and label information along distinct dimensions.

\begin{figure}[!t]
    \centering
    \includegraphics[width=0.9\linewidth]{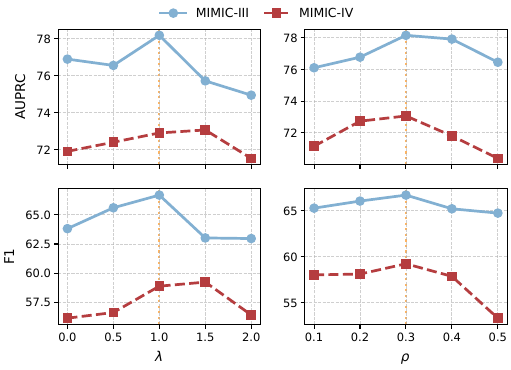}
    \caption{\textbf{Sensitivity of drug-recommendation performance to $\lambda$ and $\rho$.} Each panel sweeps one hyperparameter on MIMIC-III/IV with the other fixed at its default; the dotted vertical line marks the configuration reported in the main tables.}
    \Description{Two-by-two line plots showing sensitivity of drug-recommendation AUPRC and F1 to the lambda and rho hyperparameters on MIMIC-III and MIMIC-IV.}
    \label{fig:sensitivity}
\end{figure}

\subsection{More Quantitative Analyses}
\label{sec:quantitive}

\subsubsection{Sensitivity Analysis}
We sweep the two hyperparameters that directly control pathway balance and domain granularity: the mutual-learning weight $\lambda$ and the LCA similarity threshold $\rho$ in hierarchy pruning.
Figure~\ref{fig:sensitivity} reports drug-recommendation performance on both datasets.
The model achieves the best balance at moderate $\lambda$ ($1.0$ for MIMIC-III, $1.5$ for MIMIC-IV) and around $\rho=0.3$, while either suppressing the auxiliary objective ($\lambda=0$) or pushing it too hard ($\lambda=2$) consistently hurts both metrics; granularity behaves similarly, with very small or large $\rho$ giving over- or under-fragmented domains.
These trends confirm that the gains are not the artifact of a narrow hyperparameter setting.

\subsubsection{Runtime Analysis}
Next, we compare the training time of \model{} with two other clinical DG baselines, ManyDG and SLDG, as shown in Table~\ref{tab:runtime}.
All runtimes are measured on an NVIDIA L40S GPU.
We find that an iterative training strategy effectively balances computational overhead and performance for the entire framework.
We observe that \model{} requires a training time comparable to SLDG, since both rely on iterative parameter updates that reduce the frequency of model adjustments during inference.
ManyDG consumes more time than others, primarily because its domain assumption spawns numerous latent domains for subsequent computations.

\begin{table}[t]
    \centering
    \caption{\textbf{Running Time Comparison of Drug Recommendation (seconds per epoch).} Note that SLDG uses only the most recent admissions for prediction, following the original setting.}\label{tab:runtime}
    \begin{tabular}{lcc}
        \toprule
        \textbf{Model} & \textbf{MIMIC-III} & \textbf{MIMIC-IV} \\
        \midrule
        Base & 3.206 $\pm$ 0.1219 & 6.943 $\pm$ 0.2342 \\
        ManyDG & 5.462 $\pm$ 0.2648 & 9.215 $\pm$ 0.3781 \\
        SLDG & 4.518 $\pm$ 0.0256 & 8.439 $\pm$ 0.1329 \\
        \model{} & 4.320 $\pm$ 0.1473 & 7.871 $\pm$ 0.2415 \\
        \bottomrule
    \end{tabular}
\end{table}

\subsubsection{Effect of Training Data Size}
\begin{figure}[!t]
    \centering
    \includegraphics[width=\linewidth]{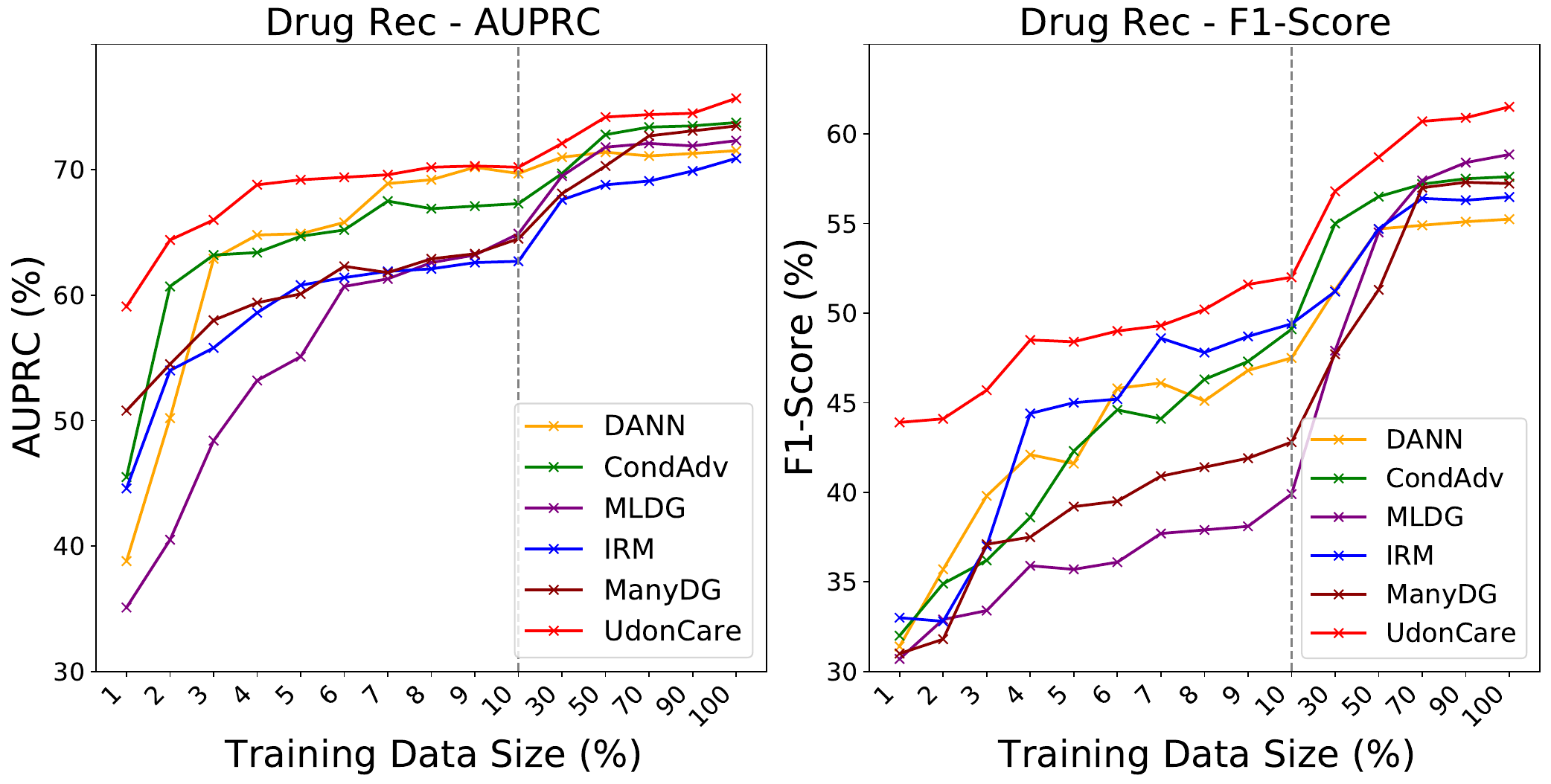}
    \caption{\textbf{Performance by Training Size.} We evaluate drug recommendation on MIMIC-IV, and the x-axis indicates \% of the entire training data. The dotted lines divide [1, 10] and [10, 100].}
    \Description{Line plot of drug-recommendation performance on MIMIC-IV as the percentage of training data increases from 1 to 100 percent.}
    \label{fig:case_b}
\end{figure}

Lastly, inspired by~\cite{yang2023manydg, jiang2024graphcare}, we investigate how the volume of training data impacts model performance by conducting a comprehensive experiment in which the training set size ranges from 1\% to 100\% to examine how well models generalize with few domain samples.
We evaluate drug recommendation on MIMIC-IV, since its complexity poses a challenging setting for prediction under varying EHR data sizes.
All reported metrics are averaged over five independent runs.
The results in Figure~\ref{fig:case_b} indicate that all models show reduced performance in both AUPRC and F1-score when labeled data are scarce, particularly below 10\% of the training set.
However, \model{} maintains a considerable edge over other baselines, suggesting that its co-training strategies effectively minimize information loss even when domain features are limited.
Notably, MLDG lacks a certain level of resilience against data limitation, likely due to unique domain assignment, which might not work for the situation that existing patients with only few admissions (less than 3) on EHR datasets.

\section{Conclusion}
In this work, we propose \model{}, a novel framework for clinical domain generalization that discovers hierarchy-grounded domains with adaptive granularity.
By pruning hierarchical concepts, our model addresses two unique challenges: the absence of domain IDs and the need for domain definitions that align with medical semantics.
Extensive evaluations on multiple EHR datasets demonstrate that \model{} outperforms existing baselines across diverse tasks, without incurring additional computational overhead.
For example, in mortality prediction, \model{} achieves up to a 12.8\% relative improvement in AUPRC over the strongest baseline.
In the future, we plan to study multi-modal clinical prediction and more general ontology-rich problems.

\begin{acks}
This research was funded by the
\grantsponsor{GS100000001}{National Science Foundation}{https://doi.org/10.13039/100000001}
under grants \grantnum{GS100000001}{IIS-2047843} and \grantnum{GS100000001}{IIS-2437621}.
Computational resources were provided through
\grantsponsor{GSACCESS}{ACCESS}{https://access-ci.org}
allocations \grantnum{GSACCESS}{MED260048} and \grantnum{GSACCESS}{MED260056}.
The views and conclusions contained in this document are those of the authors and should not be interpreted as representing the official policies, either expressed or implied, of the NSF or ACCESS.
\end{acks}

\par\vskip 1.5\baselineskip
\begin{center}
    {\LARGE \bfseries Appendix}
\end{center}

\appendix

\section{Additional Experimental Results}
\label{app:supp_main}

\begin{figure}[t]
    \centering
    \includegraphics[width=\linewidth]{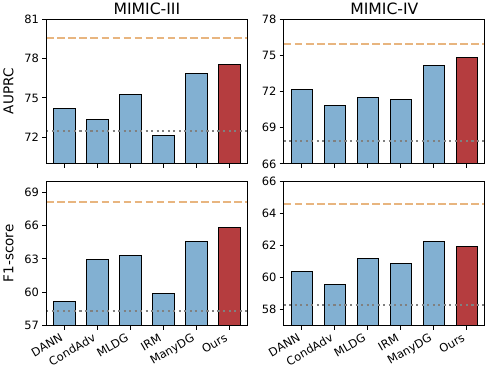}
    \caption{\textbf{The Performance of drug recommendation upon GAMENet.} Bars show the five DG baselines and \model{}; dashed and dotted horizontal lines mark Oracle and Base.}
    \Description{Two-by-two grouped bar charts comparing UdonCare with five DG baselines on AUPRC and F1-score for MIMIC-III and MIMIC-IV under the GAMENet backbone.}
    \label{fig:gamenet}
\end{figure}

\textbf{Results with GAMENet Backbone.}
We extend the main experiments by replacing the Transformer backbone with GAMENet~\citep{shang2019gamenet}, a model specifically designed for drug recommendation tasks.
We evaluate it on the drug recommendation task using both MIMIC-III and MIMIC-IV.
As shown in Figure~\ref{fig:gamenet}, the Oracle--Base gap again highlights domain shift. Standard domain generalization methods (e.g., DANN and IRM) provide limited improvements, while CondAdv and MLDG show moderate gains. ManyDG performs well on MIMIC-IV, but \model{} achieves the best overall AUPRC and F1-score. These findings confirm that \model{} generalizes effectively with a domain-specific backbone.

\textbf{Results of Cross-Institutional Distributional Shift.}
Following SLDG~\citep{wu2023iterative}, we extend the evaluation to eICU, where hospitals are categorized into four groups based on their locations: Midwest, Northeast, West, and South.
To deploy our model on this dataset, we use the preprocessing procedure provided by PyHealth.
We evaluate spatial domain shifts by treating the Midwest group as the target domain and the remaining groups as the source domain.
We use diagnosis and treatment codes to construct patient records, since medications in eICU are not encoded with standard IDs.
Table~\ref{tab:eicu_region} reports the results for readmission and diagnosis prediction. \model{} continues to outperform all baselines across both tasks.

\section{Domain Discovery with More Ontologies}
\label{app:with_more_features}

\begin{figure}[!t]
    \centering
    \includegraphics[width=\linewidth]{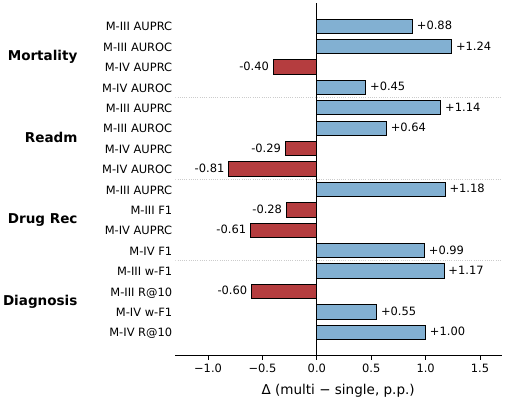}
    \caption{\textbf{Effect of adding procedure and medication ontologies on \model{}.} Each row reports the change in performance (multi-ontology $-$ single-ontology, in p.p.); blue marks improvements, red marks regressions.}
    \Description{Horizontal bar chart of per-metric delta values between UdonCare under single-ontology and multi-ontology settings, with bars color-coded by sign.}
    \label{fig:more-features}
\end{figure}

To examine whether latent domain discovery benefits from richer feature information, we further incorporate procedure and medication codes as additional keys in \model{}.
Figure~\ref{fig:more-features} reports the per-metric change relative to the single-ontology setting in Table~\ref{tab:main_result}.
We observe that gains and regressions are both small (mostly within $\pm 1$ p.p.) and inconsistent across tasks.
These findings are consistent with the tendency of additional ontologies to produce overly fine-grained domains. Patients with the same disease may be split solely by medication regimen, weakening the benefits of ontology-guided grouping.
Concretely, the pruned vocabulary increases from \textbf{438} to \textbf{1097} nodes on readmission, while the share of patients whose domain is shared by another patient falls from \textbf{89.4\%} to \textbf{53.8\%}. Consequently, patients with the same disease are more likely to be assigned to different domains solely because of distinct medication regimens.

Our reliance on the disease ontology is therefore not due to a limitation of \model{}, but reflects that diseases are a direct and broadly available source of patient information.
Almost all EHR prediction tasks use diagnosis codes as primary inputs, and diseases strongly influence domain formation.
The disease ontology can therefore serve as an appropriate basis in general clinical settings.
Rather than combining more ontologies, we encourage practitioners to select the one most appropriate for the task.
For instance, a medication ontology such as ATC may be equally or more appropriate for drug recommendation.
\model{} directly supports such substitution.
In this paper, we use the disease ontology consistently across all four tasks in our experiments.

\begin{table}[t!]
\centering
\caption{\textbf{Performance under Cross-Institutional Distributional Shift on eICU.}}
\label{tab:eicu_region}
\resizebox{\columnwidth}{!}{
\begin{tabular}{lcccc}
\toprule
\multirow{2}{*}{\textbf{Model}} 
& \multicolumn{2}{c}{\textbf{Readmission Prediction}} 
& \multicolumn{2}{c}{\textbf{Diagnosis Prediction}} \\
\cmidrule(lr){2-3} \cmidrule(lr){4-5}
& \textbf{AUPRC} & \textbf{AUROC} & \textbf{w-F$_1$} & \textbf{R@10} \\
\midrule
Oracle   & 22.74 (0.13) & 69.29 (0.13) & 63.18 (0.08) & 79.05 (0.09) \\
Base     & 11.97 (0.08) & 51.31 (0.06) & 53.89 (0.05) & 71.40 (0.05) \\
DANN     & 12.75 (0.16) & 54.28 (0.14) & 58.41 (0.18) & 75.66 (0.24) \\
CondAdv  & 13.16 (0.18) & 53.97 (0.12) & 57.04 (0.23) & 73.92 (0.17) \\
MLDG     & 14.82 (0.11) & 54.25 (0.17) & 58.13 (0.28) & 75.31 (0.21) \\
IRM      & 15.73 (0.19) & 58.31 (0.21) & 59.22 (0.16) & 76.46 (0.15) \\
PCL      & 15.14 (0.23) & 58.06 (0.24) & - & - \\
ManyDG   & 17.53 (0.17) & 60.20 (0.10) & 60.61 (0.38) & 76.94 (0.21) \\
SLDG     & 16.71 (0.21) & 59.54 (0.18) & - & - \\
\textbf{UdonCare}
         & \textbf{18.37 (0.18)} & \textbf{62.12 (0.11)}
         & \textbf{61.83 (0.14)} & \textbf{77.02 (0.11)} \\
\bottomrule
\end{tabular}
}
\end{table}

\section*{GenAI Usage Disclosure}
Generative AI tools assisted portions of this work: GPT-5.6 for manuscript polishing. All content was reviewed and verified by the authors for accuracy and compliance. 
The authors retain full responsibility for the final content of this work. No confidential or proprietary data were entered into the AI systems.

\bibliographystyle{ACM-Reference-Format}
\bibliography{main}


\begin{thebibliography}{69}


\ifx \showCODEN    \undefined \def \showCODEN     #1{\unskip}     \fi
\ifx \showISBNx    \undefined \def \showISBNx     #1{\unskip}     \fi
\ifx \showISBNxiii \undefined \def \showISBNxiii  #1{\unskip}     \fi
\ifx \showISSN     \undefined \def \showISSN      #1{\unskip}     \fi
\ifx \showLCCN     \undefined \def \showLCCN      #1{\unskip}     \fi
\ifx \shownote     \undefined \def \shownote      #1{#1}          \fi
\ifx \showarticletitle \undefined \def \showarticletitle #1{#1}   \fi
\ifx \showURL      \undefined \def \showURL       {\relax}        \fi
\providecommand\bibfield[2]{#2}
\providecommand\bibinfo[2]{#2}
\providecommand\natexlab[1]{#1}
\providecommand\showeprint[2][]{arXiv:#2}

\bibitem[Arjovsky et~al\mbox{.}(2019)]%
        {arjovsky2019invariant}
\bibfield{author}{\bibinfo{person}{Martin Arjovsky}, \bibinfo{person}{L{\'e}on
  Bottou}, \bibinfo{person}{Ishaan Gulrajani}, {and} \bibinfo{person}{David
  Lopez-Paz}.} \bibinfo{year}{2019}\natexlab{}.
\newblock \bibinfo{title}{Invariant Risk Minimization}.
\newblock
\shownote{arXiv preprint arXiv:1907.02893}.
\newblock
\urldef\tempurl%
\url{https://arxiv.org/abs/1907.02893}
\showURL{%
\tempurl}


\bibitem[Balaji et~al\mbox{.}(2018)]%
        {balaji2018metareg}
\bibfield{author}{\bibinfo{person}{Yogesh Balaji}, \bibinfo{person}{Swami
  Sankaranarayanan}, {and} \bibinfo{person}{Rama Chellappa}.}
  \bibinfo{year}{2018}\natexlab{}.
\newblock \showarticletitle{Metareg: Towards domain generalization using
  meta-regularization}.
\newblock \bibinfo{journal}{\emph{Advances in neural information processing
  systems}}  \bibinfo{volume}{31} (\bibinfo{year}{2018}),
  \bibinfo{pages}{1006--1016}.
\newblock


\bibitem[Borgwardt et~al\mbox{.}(2006)]%
        {borgwardt2006integrating}
\bibfield{author}{\bibinfo{person}{Karsten~M Borgwardt},
  \bibinfo{person}{Arthur Gretton}, \bibinfo{person}{Malte~J Rasch},
  \bibinfo{person}{Hans-Peter Kriegel}, \bibinfo{person}{Bernhard
  Sch{\"o}lkopf}, {and} \bibinfo{person}{Alex~J Smola}.}
  \bibinfo{year}{2006}\natexlab{}.
\newblock \showarticletitle{Integrating structured biological data by kernel
  maximum mean discrepancy}.
\newblock \bibinfo{journal}{\emph{Bioinformatics}} \bibinfo{volume}{22},
  \bibinfo{number}{14} (\bibinfo{year}{2006}), \bibinfo{pages}{e49--e57}.
\newblock


\bibitem[Bousmalis et~al\mbox{.}(2016)]%
        {bousmalis2016domain}
\bibfield{author}{\bibinfo{person}{Konstantinos Bousmalis},
  \bibinfo{person}{George Trigeorgis}, \bibinfo{person}{Nathan Silberman},
  \bibinfo{person}{Dilip Krishnan}, {and} \bibinfo{person}{Dumitru Erhan}.}
  \bibinfo{year}{2016}\natexlab{}.
\newblock \showarticletitle{Domain Separation Networks}. In
  \bibinfo{booktitle}{\emph{Advances in Neural Information Processing Systems
  29, 2016}}. \bibinfo{publisher}{NeurIPS}, \bibinfo{address}{Barcelona,
  Spain}, \bibinfo{pages}{343--351}.
\newblock


\bibitem[Cha et~al\mbox{.}(2021)]%
        {cha2021swad}
\bibfield{author}{\bibinfo{person}{Junbum Cha}, \bibinfo{person}{Sanghyuk
  Chun}, \bibinfo{person}{Kyungjae Lee}, \bibinfo{person}{Han-Cheol Cho},
  \bibinfo{person}{Seunghyun Park}, \bibinfo{person}{Yunsung Lee}, {and}
  \bibinfo{person}{Sungrae Park}.} \bibinfo{year}{2021}\natexlab{}.
\newblock \showarticletitle{Swad: Domain generalization by seeking flat
  minima}.
\newblock \bibinfo{journal}{\emph{Advances in Neural Information Processing
  Systems}}  \bibinfo{volume}{34} (\bibinfo{year}{2021}),
  \bibinfo{pages}{22405--22418}.
\newblock


\bibitem[Choi et~al\mbox{.}(2016)]%
        {choi2016doctor}
\bibfield{author}{\bibinfo{person}{Edward Choi}, \bibinfo{person}{Mohammad~Taha
  Bahadori}, \bibinfo{person}{Andy Schuetz}, \bibinfo{person}{Walter~F.
  Stewart}, {and} \bibinfo{person}{Jimeng Sun}.}
  \bibinfo{year}{2016}\natexlab{}.
\newblock \showarticletitle{Doctor AI: Predicting Clinical Events via Recurrent
  Neural Networks}. In \bibinfo{booktitle}{\emph{Proceedings of the 1st Machine
  Learning for Healthcare Conference}} \emph{(\bibinfo{series}{Proceedings of
  Machine Learning Research}, Vol.~\bibinfo{volume}{56})}.
  \bibinfo{publisher}{PMLR}, \bibinfo{address}{Los Angeles, CA, USA},
  \bibinfo{pages}{301--318}.
\newblock
\urldef\tempurl%
\url{https://proceedings.mlr.press/v56/Choi16.html}
\showURL{%
\tempurl}


\bibitem[Choi et~al\mbox{.}(2017)]%
        {ChoiBSSS17}
\bibfield{author}{\bibinfo{person}{Edward Choi}, \bibinfo{person}{Mohammad~Taha
  Bahadori}, \bibinfo{person}{Le Song}, \bibinfo{person}{Walter~F. Stewart},
  {and} \bibinfo{person}{Jimeng Sun}.} \bibinfo{year}{2017}\natexlab{}.
\newblock \showarticletitle{{GRAM:} Graph-based Attention Model for Healthcare
  Representation Learning}. In \bibinfo{booktitle}{\emph{Proceedings of the
  23rd {ACM} {SIGKDD} International Conference on Knowledge Discovery and Data
  Mining}}. \bibinfo{publisher}{{ACM}}, \bibinfo{address}{Halifax, NS, Canada},
  \bibinfo{pages}{787--795}.
\newblock


\bibitem[Chu et~al\mbox{.}(2022)]%
        {chu2022dna}
\bibfield{author}{\bibinfo{person}{Xu Chu}, \bibinfo{person}{Yujie Jin},
  \bibinfo{person}{Wenwu Zhu}, \bibinfo{person}{Yasha Wang},
  \bibinfo{person}{Xin Wang}, \bibinfo{person}{Shanghang Zhang}, {and}
  \bibinfo{person}{Hong Mei}.} \bibinfo{year}{2022}\natexlab{}.
\newblock \showarticletitle{Dna: Domain generalization with diversified neural
  averaging}. In \bibinfo{booktitle}{\emph{International conference on machine
  learning}}. PMLR, \bibinfo{publisher}{{PMLR}}, \bibinfo{address}{Baltimore,
  Maryland, {USA}}, \bibinfo{pages}{4010--4034}.
\newblock


\bibitem[Cui et~al\mbox{.}(2019)]%
        {cui2019deep}
\bibfield{author}{\bibinfo{person}{Runpeng Cui}, \bibinfo{person}{Hu Liu},
  {and} \bibinfo{person}{Changshui Zhang}.} \bibinfo{year}{2019}\natexlab{}.
\newblock \showarticletitle{A deep neural framework for continuous sign
  language recognition by iterative training}.
\newblock \bibinfo{journal}{\emph{IEEE Transactions on Multimedia}}
  \bibinfo{volume}{21}, \bibinfo{number}{7} (\bibinfo{year}{2019}),
  \bibinfo{pages}{1880--1891}.
\newblock


\bibitem[Ding et~al\mbox{.}(2022)]%
        {ding2022domain}
\bibfield{author}{\bibinfo{person}{Yu Ding}, \bibinfo{person}{Lei Wang},
  \bibinfo{person}{Bin Liang}, \bibinfo{person}{Shuming Liang},
  \bibinfo{person}{Yang Wang}, {and} \bibinfo{person}{Fang Chen}.}
  \bibinfo{year}{2022}\natexlab{}.
\newblock \showarticletitle{Domain Generalization by Learning and Removing
  Domain-Specific Features}. In \bibinfo{booktitle}{\emph{Advances in Neural
  Information Processing Systems}}, Vol.~\bibinfo{volume}{35}.
  \bibinfo{publisher}{Neural Information Processing Systems Foundation, Inc.},
  \bibinfo{address}{New Orleans, Louisiana, USA},
  \bibinfo{pages}{24226--24239}.
\newblock
\href{https://doi.org/10.52202/068431-1759}{doi:\nolinkurl{10.52202/068431-1759}}


\bibitem[Dou et~al\mbox{.}(2019)]%
        {dou2019domain}
\bibfield{author}{\bibinfo{person}{Qi Dou}, \bibinfo{person}{Daniel Coelho~de
  Castro}, \bibinfo{person}{Konstantinos Kamnitsas}, {and} \bibinfo{person}{Ben
  Glocker}.} \bibinfo{year}{2019}\natexlab{}.
\newblock \showarticletitle{Domain generalization via model-agnostic learning
  of semantic features}.
\newblock \bibinfo{journal}{\emph{Advances in neural information processing
  systems}}  \bibinfo{volume}{32} (\bibinfo{year}{2019}),
  \bibinfo{pages}{6447--6458}.
\newblock


\bibitem[Finn et~al\mbox{.}(2017)]%
        {finn2017model}
\bibfield{author}{\bibinfo{person}{Chelsea Finn}, \bibinfo{person}{Pieter
  Abbeel}, {and} \bibinfo{person}{Sergey Levine}.}
  \bibinfo{year}{2017}\natexlab{}.
\newblock \showarticletitle{Model-Agnostic Meta-Learning for Fast Adaptation of
  Deep Networks}. In \bibinfo{booktitle}{\emph{Proceedings of the 34th
  International Conference on Machine Learning}}
  \emph{(\bibinfo{series}{Proceedings of Machine Learning Research},
  Vol.~\bibinfo{volume}{70})}, \bibfield{editor}{\bibinfo{person}{Doina Precup}
  {and} \bibinfo{person}{Yee~Whye Teh}} (Eds.). \bibinfo{publisher}{PMLR},
  \bibinfo{address}{Sydney, Australia}, \bibinfo{pages}{1126--1135}.
\newblock
\urldef\tempurl%
\url{https://proceedings.mlr.press/v70/finn17a.html}
\showURL{%
\tempurl}


\bibitem[Ganin et~al\mbox{.}(2016)]%
        {ganin2016domain}
\bibfield{author}{\bibinfo{person}{Yaroslav Ganin}, \bibinfo{person}{Evgeniya
  Ustinova}, \bibinfo{person}{Hana Ajakan}, \bibinfo{person}{Pascal Germain},
  \bibinfo{person}{Hugo Larochelle}, \bibinfo{person}{Fran{\c{c}}ois
  Laviolette}, \bibinfo{person}{Mario Marchand}, {and} \bibinfo{person}{Victor
  Lempitsky}.} \bibinfo{year}{2016}\natexlab{}.
\newblock \showarticletitle{Domain-Adversarial Training of Neural Networks}.
\newblock \bibinfo{journal}{\emph{Journal of Machine Learning Research}}
  \bibinfo{volume}{17}, \bibinfo{number}{59} (\bibinfo{year}{2016}),
  \bibinfo{pages}{1--35}.
\newblock
\urldef\tempurl%
\url{https://www.jmlr.org/papers/v17/15-239.html}
\showURL{%
\tempurl}


\bibitem[Guo et~al\mbox{.}(2022)]%
        {guo2022evaluation}
\bibfield{author}{\bibinfo{person}{Lin~Lawrence Guo},
  \bibinfo{person}{Stephen~R Pfohl}, \bibinfo{person}{Jason Fries},
  \bibinfo{person}{Alistair~EW Johnson}, \bibinfo{person}{Jose Posada},
  \bibinfo{person}{Catherine Aftandilian}, \bibinfo{person}{Nigam Shah}, {and}
  \bibinfo{person}{Lillian Sung}.} \bibinfo{year}{2022}\natexlab{}.
\newblock \showarticletitle{Evaluation of domain generalization and adaptation
  on improving model robustness to temporal dataset shift in clinical
  medicine}.
\newblock \bibinfo{journal}{\emph{Scientific reports}} \bibinfo{volume}{12},
  \bibinfo{number}{1} (\bibinfo{year}{2022}), \bibinfo{pages}{2726}.
\newblock


\bibitem[Hai et~al\mbox{.}(2024)]%
        {hai2024domain}
\bibfield{author}{\bibinfo{person}{Ameen~Abdel Hai}, \bibinfo{person}{Mark~G
  Weiner}, \bibinfo{person}{Alice Livshits}, \bibinfo{person}{Jeremiah~R
  Brown}, \bibinfo{person}{Anuradha Paranjape}, \bibinfo{person}{Wenke Hwang},
  \bibinfo{person}{Lester~H Kirchner}, \bibinfo{person}{Nestoras Mathioudakis},
  \bibinfo{person}{Esra~Karslioglu French}, \bibinfo{person}{Zoran Obradovic},
  {et~al\mbox{.}}} \bibinfo{year}{2024}\natexlab{}.
\newblock \showarticletitle{Domain generalization for enhanced predictions of
  hospital readmission on unseen domains among patients with diabetes}.
\newblock \bibinfo{journal}{\emph{Artificial intelligence in medicine}}
  \bibinfo{volume}{158} (\bibinfo{year}{2024}), \bibinfo{pages}{103010}.
\newblock


\bibitem[Han et~al\mbox{.}(2025)]%
        {han2025black}
\bibfield{author}{\bibinfo{person}{Xiaoxue Han}, \bibinfo{person}{Pengfei Hu},
  \bibinfo{person}{Chang Lu}, \bibinfo{person}{Jun-En Ding},
  \bibinfo{person}{Feng Liu}, {and} \bibinfo{person}{Yue Ning}.}
  \bibinfo{year}{2025}\natexlab{}.
\newblock \showarticletitle{No Black Boxes: Interpretable and Interactable
  Predictive Healthcare with Knowledge-Enhanced Agentic Causal Discovery}. In
  \bibinfo{booktitle}{\emph{Findings of the Association for Computational
  Linguistics: EMNLP 2025}}. \bibinfo{publisher}{Association for Computational
  Linguistics}, \bibinfo{address}{Suzhou, China},
  \bibinfo{pages}{23415--23427}.
\newblock
\href{https://doi.org/10.18653/v1/2025.findings-emnlp.1271}{doi:\nolinkurl{10.18653/v1/2025.findings-emnlp.1271}}


\bibitem[Hu et~al\mbox{.}(2026a)]%
        {hu2026exploring}
\bibfield{author}{\bibinfo{person}{Pengfei Hu}, \bibinfo{person}{Chang Lu},
  \bibinfo{person}{Feifan Liu}, {and} \bibinfo{person}{Yue Ning}.}
  \bibinfo{year}{2026}\natexlab{a}.
\newblock \bibinfo{title}{Exploring Accurate and Transparent Domain Adaptation
  in Predictive Healthcare via Concept-Grounded Orthogonal Inference}.
\newblock \bibinfo{howpublished}{Forty-third International Conference on
  Machine Learning ({ICML})}.
\newblock
\urldef\tempurl%
\url{https://openreview.net/forum?id=KidQq3EaOe}
\showURL{%
\tempurl}


\bibitem[Hu et~al\mbox{.}(2026b)]%
        {hu2026bridging}
\bibfield{author}{\bibinfo{person}{Pengfei Hu}, \bibinfo{person}{Chang Lu},
  \bibinfo{person}{Fei Wang}, {and} \bibinfo{person}{Yue Ning}.}
  \bibinfo{year}{2026}\natexlab{b}.
\newblock \bibinfo{title}{Bridging Stepwise Lab-Informed Pretraining and
  Knowledge-Guided Learning for Diagnostic Reasoning}.
\newblock \bibinfo{howpublished}{IEEE Journal of Biomedical and Health
  Informatics}.
\newblock
\shownote{pp. 1--13}.
\newblock
\href{https://doi.org/10.1109/JBHI.2026.3670251}{doi:\nolinkurl{10.1109/JBHI.2026.3670251}}


\bibitem[Huang et~al\mbox{.}(2019)]%
        {huang2019clinicalbert}
\bibfield{author}{\bibinfo{person}{Kexin Huang}, \bibinfo{person}{Jaan
  Altosaar}, {and} \bibinfo{person}{Rajesh Ranganath}.}
  \bibinfo{year}{2019}\natexlab{}.
\newblock \bibinfo{title}{{ClinicalBERT}: Modeling Clinical Notes and
  Predicting Hospital Readmission}.
\newblock
\shownote{arXiv preprint arXiv:1904.05342}.
\newblock
\urldef\tempurl%
\url{https://arxiv.org/abs/1904.05342}
\showURL{%
\tempurl}


\bibitem[Jeon et~al\mbox{.}(2021)]%
        {jeon2021feature}
\bibfield{author}{\bibinfo{person}{Seogkyu Jeon}, \bibinfo{person}{Kibeom
  Hong}, \bibinfo{person}{Pilhyeon Lee}, \bibinfo{person}{Jewook Lee}, {and}
  \bibinfo{person}{Hyeran Byun}.} \bibinfo{year}{2021}\natexlab{}.
\newblock \showarticletitle{Feature stylization and domain-aware contrastive
  learning for domain generalization}. In \bibinfo{booktitle}{\emph{Proceedings
  of the 29th ACM International Conference on Multimedia}}.
  \bibinfo{publisher}{{ACM}}, \bibinfo{address}{Virtual Event, China},
  \bibinfo{pages}{22--31}.
\newblock


\bibitem[Jiang et~al\mbox{.}(2024)]%
        {jiang2024graphcare}
\bibfield{author}{\bibinfo{person}{Pengcheng Jiang}, \bibinfo{person}{Cao
  Xiao}, \bibinfo{person}{Adam~Richard Cross}, {and} \bibinfo{person}{Jimeng
  Sun}.} \bibinfo{year}{2024}\natexlab{}.
\newblock \bibinfo{title}{GraphCare: Enhancing Healthcare Predictions with
  Personalized Knowledge Graphs}.
\newblock \bibinfo{howpublished}{The Twelfth International Conference on
  Learning Representations ({ICLR})}.
\newblock
\urldef\tempurl%
\url{https://openreview.net/forum?id=tVTN7Zs0ml}
\showURL{%
\tempurl}


\bibitem[Johnson et~al\mbox{.}(2023)]%
        {johnson2023mimic}
\bibfield{author}{\bibinfo{person}{Alistair~EW Johnson}, \bibinfo{person}{Lucas
  Bulgarelli}, \bibinfo{person}{Lu Shen}, \bibinfo{person}{Alvin Gayles},
  \bibinfo{person}{Ayad Shammout}, \bibinfo{person}{Steven Horng},
  \bibinfo{person}{Tom~J Pollard}, \bibinfo{person}{Sicheng Hao},
  \bibinfo{person}{Benjamin Moody}, \bibinfo{person}{Brian Gow},
  {et~al\mbox{.}}} \bibinfo{year}{2023}\natexlab{}.
\newblock \showarticletitle{MIMIC-IV, a freely accessible electronic health
  record dataset}.
\newblock \bibinfo{journal}{\emph{Scientific data}} \bibinfo{volume}{10},
  \bibinfo{number}{1} (\bibinfo{year}{2023}), \bibinfo{pages}{1}.
\newblock


\bibitem[Johnson et~al\mbox{.}(2016)]%
        {johnson2016mimic}
\bibfield{author}{\bibinfo{person}{Alistair~EW Johnson}, \bibinfo{person}{Tom~J
  Pollard}, \bibinfo{person}{Lu Shen}, \bibinfo{person}{Li-wei~H Lehman},
  \bibinfo{person}{Mengling Feng}, \bibinfo{person}{Mohammad Ghassemi},
  \bibinfo{person}{Benjamin Moody}, \bibinfo{person}{Peter Szolovits},
  \bibinfo{person}{Leo Anthony~Celi}, {and} \bibinfo{person}{Roger~G Mark}.}
  \bibinfo{year}{2016}\natexlab{}.
\newblock \showarticletitle{MIMIC-III, a freely accessible critical care
  database}.
\newblock \bibinfo{journal}{\emph{Scientific data}} \bibinfo{volume}{3},
  \bibinfo{number}{1} (\bibinfo{year}{2016}), \bibinfo{pages}{1--9}.
\newblock


\bibitem[Johnson(1967)]%
        {johnson1967hierarchical}
\bibfield{author}{\bibinfo{person}{Stephen~C Johnson}.}
  \bibinfo{year}{1967}\natexlab{}.
\newblock \showarticletitle{Hierarchical clustering schemes}.
\newblock \bibinfo{journal}{\emph{Psychometrika}} \bibinfo{volume}{32},
  \bibinfo{number}{3} (\bibinfo{year}{1967}), \bibinfo{pages}{241--254}.
\newblock


\bibitem[Kim et~al\mbox{.}(2021)]%
        {kim2021selfreg}
\bibfield{author}{\bibinfo{person}{Daehee Kim}, \bibinfo{person}{Youngjun Yoo},
  \bibinfo{person}{Seunghyun Park}, \bibinfo{person}{Jinkyu Kim}, {and}
  \bibinfo{person}{Jaekoo Lee}.} \bibinfo{year}{2021}\natexlab{}.
\newblock \showarticletitle{Selfreg: Self-supervised contrastive regularization
  for domain generalization}. In \bibinfo{booktitle}{\emph{Proceedings of the
  IEEE/CVF International Conference on Computer Vision}}.
  \bibinfo{publisher}{{IEEE}}, \bibinfo{address}{Montreal, QC, Canada},
  \bibinfo{pages}{9619--9628}.
\newblock


\bibitem[Koh et~al\mbox{.}(2021)]%
        {koh2021wilds}
\bibfield{author}{\bibinfo{person}{Pang~Wei Koh}, \bibinfo{person}{Shiori
  Sagawa}, \bibinfo{person}{Henrik Marklund}, \bibinfo{person}{Sang~Michael
  Xie}, \bibinfo{person}{Marvin Zhang}, \bibinfo{person}{Akshay Balsubramani},
  \bibinfo{person}{Weihua Hu}, \bibinfo{person}{Michihiro Yasunaga},
  \bibinfo{person}{Richard~Lanas Phillips}, \bibinfo{person}{Irena Gao},
  {et~al\mbox{.}}} \bibinfo{year}{2021}\natexlab{}.
\newblock \showarticletitle{Wilds: A benchmark of in-the-wild distribution
  shifts}. In \bibinfo{booktitle}{\emph{International conference on machine
  learning}}. \bibinfo{publisher}{{PMLR}}, \bibinfo{address}{Virtual Event},
  \bibinfo{pages}{5637--5664}.
\newblock


\bibitem[Li et~al\mbox{.}(2018b)]%
        {li2018learning}
\bibfield{author}{\bibinfo{person}{Da Li}, \bibinfo{person}{Yongxin Yang},
  \bibinfo{person}{Yi{-}Zhe Song}, {and} \bibinfo{person}{Timothy~M.
  Hospedales}.} \bibinfo{year}{2018}\natexlab{b}.
\newblock \showarticletitle{Learning to Generalize: Meta-Learning for Domain
  Generalization}. In \bibinfo{booktitle}{\emph{Proceedings of the
  Thirty-Second {AAAI} Conference on Artificial Intelligence}}.
  \bibinfo{publisher}{{AAAI} Press}, \bibinfo{address}{New Orleans, Louisiana,
  USA}, \bibinfo{pages}{3490--3497}.
\newblock


\bibitem[Li et~al\mbox{.}(2020)]%
        {li2020domain}
\bibfield{author}{\bibinfo{person}{Haoliang Li}, \bibinfo{person}{YuFei Wang},
  \bibinfo{person}{Renjie Wan}, \bibinfo{person}{Shiqi Wang},
  \bibinfo{person}{Tie-Qiang Li}, {and} \bibinfo{person}{Alex Kot}.}
  \bibinfo{year}{2020}\natexlab{}.
\newblock \showarticletitle{Domain generalization for medical imaging
  classification with linear-dependency regularization}.
\newblock \bibinfo{journal}{\emph{Advances in neural information processing
  systems}}  \bibinfo{volume}{33} (\bibinfo{year}{2020}),
  \bibinfo{pages}{3118--3129}.
\newblock


\bibitem[Li et~al\mbox{.}(2018a)]%
        {li2018deep}
\bibfield{author}{\bibinfo{person}{Ya Li}, \bibinfo{person}{Xinmei Tian},
  \bibinfo{person}{Mingming Gong}, \bibinfo{person}{Yajing Liu},
  \bibinfo{person}{Tongliang Liu}, \bibinfo{person}{Kun Zhang}, {and}
  \bibinfo{person}{Dacheng Tao}.} \bibinfo{year}{2018}\natexlab{a}.
\newblock \showarticletitle{Deep Domain Generalization via Conditional
  Invariant Adversarial Networks}. In \bibinfo{booktitle}{\emph{Computer Vision
  -- ECCV 2018}} \emph{(\bibinfo{series}{Lecture Notes in Computer Science},
  Vol.~\bibinfo{volume}{11219})}. \bibinfo{publisher}{Springer},
  \bibinfo{address}{Munich, Germany}, \bibinfo{pages}{647--663}.
\newblock
\href{https://doi.org/10.1007/978-3-030-01267-0_38}{doi:\nolinkurl{10.1007/978-3-030-01267-0_38}}


\bibitem[Lin et~al\mbox{.}(2026)]%
        {lin2026medcausalx}
\bibfield{author}{\bibinfo{person}{Jianxin Lin}, \bibinfo{person}{Chunzheng
  Zhu}, \bibinfo{person}{Peter~J Kneuertz}, \bibinfo{person}{Yunfei Bai}, {and}
  \bibinfo{person}{Yuan Xue}.} \bibinfo{year}{2026}\natexlab{}.
\newblock \bibinfo{title}{When Models Learn to Ask Why: Adaptive Causal
  Reasoning for Trustworthy Medical Vision-Language Models}.
\newblock
\shownote{arXiv preprint arXiv:2603.23085}.
\newblock
\urldef\tempurl%
\url{https://arxiv.org/abs/2603.23085}
\showURL{%
\tempurl}


\bibitem[Liu et~al\mbox{.}(2021a)]%
        {liu2021domain}
\bibfield{author}{\bibinfo{person}{Chang Liu}, \bibinfo{person}{Lichen Wang},
  \bibinfo{person}{Kai Li}, {and} \bibinfo{person}{Yun Fu}.}
  \bibinfo{year}{2021}\natexlab{a}.
\newblock \showarticletitle{Domain generalization via feature variation
  decorrelation}. In \bibinfo{booktitle}{\emph{Proceedings of the 29th ACM
  International Conference on Multimedia}}. \bibinfo{publisher}{{ACM}},
  \bibinfo{address}{Virtual Event, China}, \bibinfo{pages}{1683--1691}.
\newblock


\bibitem[Liu et~al\mbox{.}(2021b)]%
        {liu2021learning}
\bibfield{author}{\bibinfo{person}{Yunan Liu}, \bibinfo{person}{Shanshan
  Zhang}, \bibinfo{person}{Yang Li}, {and} \bibinfo{person}{Jian Yang}.}
  \bibinfo{year}{2021}\natexlab{b}.
\newblock \showarticletitle{Learning to adapt via latent domains for adaptive
  semantic segmentation}.
\newblock \bibinfo{journal}{\emph{Advances in Neural Information Processing
  Systems}}  \bibinfo{volume}{34} (\bibinfo{year}{2021}),
  \bibinfo{pages}{1167--1178}.
\newblock


\bibitem[Lu et~al\mbox{.}(2021)]%
        {LuRCKN21}
\bibfield{author}{\bibinfo{person}{Chang Lu}, \bibinfo{person}{Chandan~K.
  Reddy}, \bibinfo{person}{Prithwish Chakraborty}, \bibinfo{person}{Samantha
  Kleinberg}, {and} \bibinfo{person}{Yue Ning}.}
  \bibinfo{year}{2021}\natexlab{}.
\newblock \showarticletitle{Collaborative Graph Learning with Auxiliary Text
  for Temporal Event Prediction in Healthcare}. In
  \bibinfo{booktitle}{\emph{Proceedings of the Thirtieth International Joint
  Conference on Artificial Intelligence}}. \bibinfo{publisher}{ijcai.org},
  \bibinfo{address}{Montreal, Canada}, \bibinfo{pages}{3529--3535}.
\newblock


\bibitem[Matsuura and Harada(2020)]%
        {matsuura2020domain}
\bibfield{author}{\bibinfo{person}{Toshihiko Matsuura} {and}
  \bibinfo{person}{Tatsuya Harada}.} \bibinfo{year}{2020}\natexlab{}.
\newblock \showarticletitle{Domain generalization using a mixture of multiple
  latent domains}. In \bibinfo{booktitle}{\emph{Proceedings of the AAAI
  conference on artificial intelligence}}, Vol.~\bibinfo{volume}{34}.
  \bibinfo{publisher}{{AAAI} Press}, \bibinfo{address}{New York, NY, USA},
  \bibinfo{pages}{11749--11756}.
\newblock


\bibitem[McInnes et~al\mbox{.}(2018)]%
        {mcinnes2018umap}
\bibfield{author}{\bibinfo{person}{Leland McInnes}, \bibinfo{person}{John
  Healy}, \bibinfo{person}{Nathaniel Saul}, {and} \bibinfo{person}{Lukas
  Gro{\ss}berger}.} \bibinfo{year}{2018}\natexlab{}.
\newblock \showarticletitle{{UMAP}: Uniform Manifold Approximation and
  Projection}.
\newblock \bibinfo{journal}{\emph{Journal of Open Source Software}}
  \bibinfo{volume}{3}, \bibinfo{number}{29} (\bibinfo{year}{2018}),
  \bibinfo{pages}{861}.
\newblock
\href{https://doi.org/10.21105/joss.00861}{doi:\nolinkurl{10.21105/joss.00861}}


\bibitem[Muandet et~al\mbox{.}(2013)]%
        {muandet2013domain}
\bibfield{author}{\bibinfo{person}{Krikamol Muandet}, \bibinfo{person}{David
  Balduzzi}, {and} \bibinfo{person}{Bernhard Sch{\"o}lkopf}.}
  \bibinfo{year}{2013}\natexlab{}.
\newblock \showarticletitle{Domain generalization via invariant feature
  representation}. In \bibinfo{booktitle}{\emph{International Conference on
  Machine Learning}}. PMLR, \bibinfo{publisher}{JMLR.org},
  \bibinfo{address}{Atlanta, GA, USA}, \bibinfo{pages}{10--18}.
\newblock


\bibitem[Nahler(2009)]%
        {nahler2009anatomical}
\bibfield{author}{\bibinfo{person}{Gerhard Nahler}.}
  \bibinfo{year}{2009}\natexlab{}.
\newblock \showarticletitle{Anatomical Therapeutic Chemical Classification
  System (ATC)}.
\newblock In \bibinfo{booktitle}{\emph{Dictionary of Pharmaceutical Medicine}}.
  \bibinfo{publisher}{Springer Vienna}, \bibinfo{address}{Vienna},
  \bibinfo{pages}{8--8}.
\newblock
\href{https://doi.org/10.1007/978-3-211-89836-9_64}{doi:\nolinkurl{10.1007/978-3-211-89836-9_64}}


\bibitem[Perone et~al\mbox{.}(2019)]%
        {perone2019unsupervised}
\bibfield{author}{\bibinfo{person}{Christian~S Perone}, \bibinfo{person}{Pedro
  Ballester}, \bibinfo{person}{Rodrigo~C Barros}, {and} \bibinfo{person}{Julien
  Cohen-Adad}.} \bibinfo{year}{2019}\natexlab{}.
\newblock \showarticletitle{Unsupervised domain adaptation for medical imaging
  segmentation with self-ensembling}.
\newblock \bibinfo{journal}{\emph{NeuroImage}}  \bibinfo{volume}{194}
  (\bibinfo{year}{2019}), \bibinfo{pages}{1--11}.
\newblock


\bibitem[Poulain and Beheshti(2024)]%
        {poulain2024graph}
\bibfield{author}{\bibinfo{person}{Raphael Poulain} {and}
  \bibinfo{person}{Rahmatollah Beheshti}.} \bibinfo{year}{2024}\natexlab{}.
\newblock \bibinfo{title}{Graph Transformers on EHRs: Better Representation
  Improves Downstream Performance}.
\newblock \bibinfo{howpublished}{The Twelfth International Conference on
  Learning Representations ({ICLR})}.
\newblock
\urldef\tempurl%
\url{https://openreview.net/forum?id=pe0Vdv7rsL}
\showURL{%
\tempurl}


\bibitem[Reps et~al\mbox{.}(2022)]%
        {reps2022learning}
\bibfield{author}{\bibinfo{person}{Jenna~Marie Reps}, \bibinfo{person}{Ross~D
  Williams}, \bibinfo{person}{Martijn~J Schuemie}, \bibinfo{person}{Patrick~B
  Ryan}, {and} \bibinfo{person}{Peter~R Rijnbeek}.}
  \bibinfo{year}{2022}\natexlab{}.
\newblock \showarticletitle{Learning patient-level prediction models across
  multiple healthcare databases: evaluation of ensembles for increasing model
  transportability}.
\newblock \bibinfo{journal}{\emph{BMC medical informatics and decision making}}
  \bibinfo{volume}{22}, \bibinfo{number}{1} (\bibinfo{year}{2022}),
  \bibinfo{pages}{142}.
\newblock


\bibitem[Shahapure and Nicholas(2020)]%
        {shahapure2020cluster}
\bibfield{author}{\bibinfo{person}{Ketan~Rajshekhar Shahapure} {and}
  \bibinfo{person}{Charles Nicholas}.} \bibinfo{year}{2020}\natexlab{}.
\newblock \showarticletitle{Cluster Quality Analysis Using Silhouette Score}.
  In \bibinfo{booktitle}{\emph{2020 IEEE 7th International Conference on Data
  Science and Advanced Analytics (DSAA)}}. \bibinfo{publisher}{IEEE},
  \bibinfo{address}{Sydney, NSW, Australia}, \bibinfo{pages}{747--748}.
\newblock
\href{https://doi.org/10.1109/DSAA49011.2020.00096}{doi:\nolinkurl{10.1109/DSAA49011.2020.00096}}


\bibitem[Shang et~al\mbox{.}(2019a)]%
        {ShangMXS19}
\bibfield{author}{\bibinfo{person}{Junyuan Shang}, \bibinfo{person}{Tengfei
  Ma}, \bibinfo{person}{Cao Xiao}, {and} \bibinfo{person}{Jimeng Sun}.}
  \bibinfo{year}{2019}\natexlab{a}.
\newblock \showarticletitle{Pre-training of Graph Augmented Transformers for
  Medication Recommendation}. In \bibinfo{booktitle}{\emph{Proceedings of the
  Twenty-Eighth International Joint Conference on Artificial Intelligence}}.
  \bibinfo{publisher}{ijcai.org}, \bibinfo{address}{Macao, China},
  \bibinfo{pages}{5953--5959}.
\newblock


\bibitem[Shang et~al\mbox{.}(2019b)]%
        {shang2019gamenet}
\bibfield{author}{\bibinfo{person}{Junyuan Shang}, \bibinfo{person}{Cao Xiao},
  \bibinfo{person}{Tengfei Ma}, \bibinfo{person}{Hongyan Li}, {and}
  \bibinfo{person}{Jimeng Sun}.} \bibinfo{year}{2019}\natexlab{b}.
\newblock \showarticletitle{Gamenet: Graph augmented memory networks for
  recommending medication combination}. In
  \bibinfo{booktitle}{\emph{proceedings of the AAAI Conference on Artificial
  Intelligence}}, Vol.~\bibinfo{volume}{33}. \bibinfo{publisher}{{AAAI} Press},
  \bibinfo{address}{Honolulu, Hawaii, USA}, \bibinfo{pages}{1126--1133}.
\newblock


\bibitem[Shen et~al\mbox{.}(2022)]%
        {shen2022connect}
\bibfield{author}{\bibinfo{person}{Kendrick Shen}, \bibinfo{person}{Robbie~M.
  Jones}, \bibinfo{person}{Ananya Kumar}, \bibinfo{person}{Sang~Michael Xie},
  \bibinfo{person}{Jeff~Z. HaoChen}, \bibinfo{person}{Tengyu Ma}, {and}
  \bibinfo{person}{Percy Liang}.} \bibinfo{year}{2022}\natexlab{}.
\newblock \showarticletitle{Connect, Not Collapse: Explaining Contrastive
  Learning for Unsupervised Domain Adaptation}. In
  \bibinfo{booktitle}{\emph{International Conference on Machine Learning,
  {ICML} 2022}}, Vol.~\bibinfo{volume}{162}. \bibinfo{publisher}{{PMLR}},
  \bibinfo{address}{Baltimore, Maryland, {USA}}, \bibinfo{pages}{19847--19878}.
\newblock


\bibitem[Sofiiuk et~al\mbox{.}(2022)]%
        {sofiiuk2022reviving}
\bibfield{author}{\bibinfo{person}{Konstantin Sofiiuk},
  \bibinfo{person}{Ilya~A. Petrov}, {and} \bibinfo{person}{Anton Konushin}.}
  \bibinfo{year}{2022}\natexlab{}.
\newblock \showarticletitle{Reviving Iterative Training with Mask Guidance for
  Interactive Segmentation}. In \bibinfo{booktitle}{\emph{2022 IEEE
  International Conference on Image Processing (ICIP)}}.
  \bibinfo{publisher}{IEEE}, \bibinfo{address}{Bordeaux, France},
  \bibinfo{pages}{3141--3145}.
\newblock
\href{https://doi.org/10.1109/ICIP46576.2022.9897365}{doi:\nolinkurl{10.1109/ICIP46576.2022.9897365}}


\bibitem[Song and Lu(2015)]%
        {song2015decision}
\bibfield{author}{\bibinfo{person}{Yan-Yan Song} {and} \bibinfo{person}{Ying
  Lu}.} \bibinfo{year}{2015}\natexlab{}.
\newblock \showarticletitle{Decision Tree Methods: Applications for
  Classification and Prediction}.
\newblock \bibinfo{journal}{\emph{Shanghai Archives of Psychiatry}}
  \bibinfo{volume}{27}, \bibinfo{number}{2} (\bibinfo{year}{2015}),
  \bibinfo{pages}{130--135}.
\newblock
\href{https://doi.org/10.11919/j.issn.1002-0829.215044}{doi:\nolinkurl{10.11919/j.issn.1002-0829.215044}}


\bibitem[Sutskever et~al\mbox{.}(2014)]%
        {sutskever2014sequence}
\bibfield{author}{\bibinfo{person}{Ilya Sutskever}, \bibinfo{person}{Oriol
  Vinyals}, {and} \bibinfo{person}{Quoc~V Le}.}
  \bibinfo{year}{2014}\natexlab{}.
\newblock \showarticletitle{Sequence to sequence learning with neural
  networks}.
\newblock \bibinfo{journal}{\emph{Advances in neural information processing
  systems}}  \bibinfo{volume}{27} (\bibinfo{year}{2014}),
  \bibinfo{pages}{3104--3112}.
\newblock


\bibitem[Tong et~al\mbox{.}(2025a)]%
        {tong2025renaissance}
\bibfield{author}{\bibinfo{person}{Ran Tong}, \bibinfo{person}{Jiaqi Liu},
  \bibinfo{person}{Su Liu}, \bibinfo{person}{Xin Hu}, {and}
  \bibinfo{person}{Lanruo Wang}.} \bibinfo{year}{2025}\natexlab{a}.
\newblock \bibinfo{title}{Renaissance of RNNs in Streaming Clinical Time
  Series: Compact Recurrence Remains Competitive with Transformers}.
\newblock
\shownote{arXiv preprint arXiv:2510.16677}.
\newblock
\urldef\tempurl%
\url{https://arxiv.org/abs/2510.16677}
\showURL{%
\tempurl}


\bibitem[Tong et~al\mbox{.}(2025b)]%
        {tong2025does}
\bibfield{author}{\bibinfo{person}{Ran Tong}, \bibinfo{person}{Jiaqi Liu},
  \bibinfo{person}{Su Liu}, \bibinfo{person}{Jiexi Xu}, \bibinfo{person}{Lanruo
  Wang}, {and} \bibinfo{person}{Tong Wang}.} \bibinfo{year}{2025}\natexlab{b}.
\newblock \bibinfo{title}{Does Bigger Mean Better? Comparitive Analysis of CNNs
  and Biomedical Vision Language Modles in Medical Diagnosis}.
\newblock
\shownote{arXiv preprint arXiv:2510.00411}.
\newblock
\urldef\tempurl%
\url{https://arxiv.org/abs/2510.00411}
\showURL{%
\tempurl}


\bibitem[Van~Erven and Harremos(2014)]%
        {van2014renyi}
\bibfield{author}{\bibinfo{person}{Tim Van~Erven} {and} \bibinfo{person}{Peter
  Harremos}.} \bibinfo{year}{2014}\natexlab{}.
\newblock \showarticletitle{R{\'e}nyi divergence and Kullback-Leibler
  divergence}.
\newblock \bibinfo{journal}{\emph{IEEE Transactions on Information Theory}}
  \bibinfo{volume}{60}, \bibinfo{number}{7} (\bibinfo{year}{2014}),
  \bibinfo{pages}{3797--3820}.
\newblock


\bibitem[Vaswani et~al\mbox{.}(2017)]%
        {vaswani2023attentionneed}
\bibfield{author}{\bibinfo{person}{Ashish Vaswani}, \bibinfo{person}{Noam
  Shazeer}, \bibinfo{person}{Niki Parmar}, \bibinfo{person}{Jakob Uszkoreit},
  \bibinfo{person}{Llion Jones}, \bibinfo{person}{Aidan~N. Gomez},
  \bibinfo{person}{{\L}ukasz Kaiser}, {and} \bibinfo{person}{Illia
  Polosukhin}.} \bibinfo{year}{2017}\natexlab{}.
\newblock \showarticletitle{Attention Is All You Need}. In
  \bibinfo{booktitle}{\emph{Advances in Neural Information Processing
  Systems}}, Vol.~\bibinfo{volume}{30}. \bibinfo{publisher}{Curran Associates,
  Inc.}, \bibinfo{address}{Long Beach, CA, USA}, \bibinfo{pages}{5998--6008}.
\newblock
\urldef\tempurl%
\url{https://proceedings.neurips.cc/paper/7181-attention-is-all-you-need}
\showURL{%
\tempurl}


\bibitem[Wang et~al\mbox{.}(2022)]%
        {wang2022variational}
\bibfield{author}{\bibinfo{person}{Yufei Wang}, \bibinfo{person}{Haoliang Li},
  \bibinfo{person}{Hao Cheng}, \bibinfo{person}{Bihan Wen},
  \bibinfo{person}{Lap-Pui Chau}, {and} \bibinfo{person}{Alex~C. Kot}.}
  \bibinfo{year}{2022}\natexlab{}.
\newblock \bibinfo{title}{Variational Disentanglement for Domain
  Generalization}.
\newblock \bibinfo{howpublished}{Transactions on Machine Learning Research}.
\newblock
\shownote{ISSN 2835-8856}.
\newblock
\urldef\tempurl%
\url{https://openreview.net/forum?id=fudOtITMIZ}
\showURL{%
\tempurl}


\bibitem[{World Health Organization}(1988)]%
        {world1988international}
\bibfield{author}{\bibinfo{person}{{World Health Organization}}.}
  \bibinfo{year}{1988}\natexlab{}.
\newblock \showarticletitle{International Classification of Diseases---Ninth
  Revision ({ICD}-9)}.
\newblock \bibinfo{journal}{\emph{Weekly Epidemiological Record}}
  \bibinfo{volume}{63}, \bibinfo{number}{45} (\bibinfo{year}{1988}),
  \bibinfo{pages}{343--344}.
\newblock
\urldef\tempurl%
\url{https://iris.who.int/handle/10665/226922}
\showURL{%
\tempurl}


\bibitem[Wu et~al\mbox{.}(2023)]%
        {wu2023iterative}
\bibfield{author}{\bibinfo{person}{Zhenbang Wu}, \bibinfo{person}{Huaxiu Yao},
  \bibinfo{person}{David Liebovitz}, {and} \bibinfo{person}{Jimeng Sun}.}
  \bibinfo{year}{2023}\natexlab{}.
\newblock \showarticletitle{An iterative self-learning framework for medical
  domain generalization}.
\newblock \bibinfo{journal}{\emph{Advances in Neural Information Processing
  Systems}}  \bibinfo{volume}{36} (\bibinfo{year}{2023}),
  \bibinfo{pages}{54833--54854}.
\newblock


\bibitem[Xu et~al\mbox{.}(2024)]%
        {xu2024ram}
\bibfield{author}{\bibinfo{person}{Ran Xu}, \bibinfo{person}{Wenqi Shi},
  \bibinfo{person}{Yue Yu}, \bibinfo{person}{Yuchen Zhuang},
  \bibinfo{person}{Bowen Jin}, \bibinfo{person}{May~Dongmei Wang},
  \bibinfo{person}{Joyce Ho}, {and} \bibinfo{person}{Carl Yang}.}
  \bibinfo{year}{2024}\natexlab{}.
\newblock \showarticletitle{{RAM}-{EHR}: Retrieval Augmentation Meets Clinical
  Predictions on Electronic Health Records}. In
  \bibinfo{booktitle}{\emph{Proceedings of the 62nd Annual Meeting of the
  Association for Computational Linguistics (Volume 2: Short Papers)}}.
  \bibinfo{publisher}{Association for Computational Linguistics},
  \bibinfo{address}{Bangkok, Thailand}, \bibinfo{pages}{754--765}.
\newblock
\href{https://doi.org/10.18653/v1/2024.acl-short.68}{doi:\nolinkurl{10.18653/v1/2024.acl-short.68}}


\bibitem[Xu et~al\mbox{.}(2023)]%
        {xu2023seqcare}
\bibfield{author}{\bibinfo{person}{Yongxin Xu}, \bibinfo{person}{Xu Chu},
  \bibinfo{person}{Kai Yang}, \bibinfo{person}{Zhiyuan Wang},
  \bibinfo{person}{Peinie Zou}, \bibinfo{person}{Hongxin Ding},
  \bibinfo{person}{Junfeng Zhao}, \bibinfo{person}{Yasha Wang}, {and}
  \bibinfo{person}{Bing Xie}.} \bibinfo{year}{2023}\natexlab{}.
\newblock \showarticletitle{SeqCare: Sequential Training with External Medical
  Knowledge Graph for Diagnosis Prediction in Healthcare Data}. In
  \bibinfo{booktitle}{\emph{Proceedings of the ACM Web Conference 2023}}.
  \bibinfo{publisher}{ACM}, \bibinfo{address}{Austin, TX, USA},
  \bibinfo{pages}{2819--2830}.
\newblock
\href{https://doi.org/10.1145/3543507.3583543}{doi:\nolinkurl{10.1145/3543507.3583543}}


\bibitem[Yang et~al\mbox{.}(2023a)]%
        {yang2023manydg}
\bibfield{author}{\bibinfo{person}{Chaoqi Yang}, \bibinfo{person}{M.~Brandon
  Westover}, {and} \bibinfo{person}{Jimeng Sun}.}
  \bibinfo{year}{2023}\natexlab{a}.
\newblock \bibinfo{title}{ManyDG: Many-Domain Generalization for Healthcare
  Applications}.
\newblock \bibinfo{howpublished}{The Eleventh International Conference on
  Learning Representations ({ICLR})}.
\newblock
\urldef\tempurl%
\url{https://openreview.net/forum?id=lcSfirnflpW}
\showURL{%
\tempurl}


\bibitem[Yang et~al\mbox{.}(2023b)]%
        {yang2023pyhealth}
\bibfield{author}{\bibinfo{person}{Chaoqi Yang}, \bibinfo{person}{Zhenbang Wu},
  \bibinfo{person}{Patrick Jiang}, \bibinfo{person}{Zhen Lin},
  \bibinfo{person}{Junyi Gao}, \bibinfo{person}{Benjamin~P Danek}, {and}
  \bibinfo{person}{Jimeng Sun}.} \bibinfo{year}{2023}\natexlab{b}.
\newblock \showarticletitle{Pyhealth: A deep learning toolkit for healthcare
  applications}. In \bibinfo{booktitle}{\emph{Proceedings of the 29th ACM
  SIGKDD Conference on Knowledge Discovery and Data Mining}}.
  \bibinfo{publisher}{{ACM}}, \bibinfo{address}{Long Beach, CA, USA},
  \bibinfo{pages}{5788--5789}.
\newblock


\bibitem[Yang et~al\mbox{.}(2024)]%
        {yang2024mplite}
\bibfield{author}{\bibinfo{person}{Eric Yang}, \bibinfo{person}{Pengfei Hu},
  \bibinfo{person}{Xiaoxue Han}, {and} \bibinfo{person}{Yue Ning}.}
  \bibinfo{year}{2024}\natexlab{}.
\newblock \showarticletitle{{MPLite}: Multi-Aspect Pretraining for Mining
  Clinical Health Records}. In \bibinfo{booktitle}{\emph{2024 IEEE
  International Conference on Big Data (BigData)}}. \bibinfo{publisher}{IEEE},
  \bibinfo{address}{Washington, DC, USA}, \bibinfo{pages}{5096--5102}.
\newblock
\href{https://doi.org/10.1109/BigData62323.2024.10825511}{doi:\nolinkurl{10.1109/BigData62323.2024.10825511}}


\bibitem[Yao et~al\mbox{.}(2022)]%
        {yao2022pcl}
\bibfield{author}{\bibinfo{person}{Xufeng Yao}, \bibinfo{person}{Yang Bai},
  \bibinfo{person}{Xinyun Zhang}, \bibinfo{person}{Yuechen Zhang},
  \bibinfo{person}{Qi Sun}, \bibinfo{person}{Ran Chen}, \bibinfo{person}{Ruiyu
  Li}, {and} \bibinfo{person}{Bei Yu}.} \bibinfo{year}{2022}\natexlab{}.
\newblock \showarticletitle{Pcl: Proxy-based contrastive learning for domain
  generalization}. In \bibinfo{booktitle}{\emph{Proceedings of the IEEE/CVF
  Conference on Computer Vision and Pattern Recognition}}.
  \bibinfo{publisher}{IEEE}, \bibinfo{address}{New Orleans, LA, USA},
  \bibinfo{pages}{7097--7107}.
\newblock


\bibitem[Zhang et~al\mbox{.}(2021a)]%
        {zhang2021empirical}
\bibfield{author}{\bibinfo{person}{Haoran Zhang}, \bibinfo{person}{Natalie
  Dullerud}, \bibinfo{person}{Laleh Seyyed-Kalantari}, \bibinfo{person}{Quaid
  Morris}, \bibinfo{person}{Shalmali Joshi}, {and} \bibinfo{person}{Marzyeh
  Ghassemi}.} \bibinfo{year}{2021}\natexlab{a}.
\newblock \showarticletitle{An empirical framework for domain generalization in
  clinical settings}. In \bibinfo{booktitle}{\emph{Proceedings of the
  conference on health, inference, and learning}}. \bibinfo{publisher}{{ACM}},
  \bibinfo{address}{Virtual Event, USA}, \bibinfo{pages}{279--290}.
\newblock


\bibitem[Zhang et~al\mbox{.}(2022b)]%
        {zhang2022towards}
\bibfield{author}{\bibinfo{person}{Hanlin Zhang}, \bibinfo{person}{Yi-Fan
  Zhang}, \bibinfo{person}{Weiyang Liu}, \bibinfo{person}{Adrian Weller},
  \bibinfo{person}{Bernhard Sch{\"o}lkopf}, {and} \bibinfo{person}{Eric~P
  Xing}.} \bibinfo{year}{2022}\natexlab{b}.
\newblock \showarticletitle{Towards principled disentanglement for domain
  generalization}. In \bibinfo{booktitle}{\emph{CVPR}}.
  \bibinfo{publisher}{{IEEE}}, \bibinfo{address}{New Orleans, LA, USA},
  \bibinfo{pages}{8024--8034}.
\newblock


\bibitem[Zhang et~al\mbox{.}(2021b)]%
        {zhang2021adaptive}
\bibfield{author}{\bibinfo{person}{Marvin Zhang}, \bibinfo{person}{Henrik
  Marklund}, \bibinfo{person}{Nikita Dhawan}, \bibinfo{person}{Abhishek Gupta},
  \bibinfo{person}{Sergey Levine}, {and} \bibinfo{person}{Chelsea Finn}.}
  \bibinfo{year}{2021}\natexlab{b}.
\newblock \showarticletitle{Adaptive risk minimization: Learning to adapt to
  domain shift}.
\newblock \bibinfo{journal}{\emph{Advances in Neural Information Processing
  Systems}}  \bibinfo{volume}{34} (\bibinfo{year}{2021}),
  \bibinfo{pages}{23664--23678}.
\newblock


\bibitem[Zhang et~al\mbox{.}(2022a)]%
        {zhang2022adadiag}
\bibfield{author}{\bibinfo{person}{Tianran Zhang}, \bibinfo{person}{Muhao
  Chen}, {and} \bibinfo{person}{Alex~AT Bui}.}
  \bibinfo{year}{2022}\natexlab{a}.
\newblock \showarticletitle{AdaDiag: Adversarial domain adaptation of
  diagnostic prediction with clinical event sequences}.
\newblock \bibinfo{journal}{\emph{Journal of biomedical informatics}}
  \bibinfo{volume}{134} (\bibinfo{year}{2022}), \bibinfo{pages}{104168}.
\newblock


\bibitem[Zhao et~al\mbox{.}(2020)]%
        {zhao2020domain}
\bibfield{author}{\bibinfo{person}{Shanshan Zhao}, \bibinfo{person}{Mingming
  Gong}, \bibinfo{person}{Tongliang Liu}, \bibinfo{person}{Huan Fu}, {and}
  \bibinfo{person}{Dacheng Tao}.} \bibinfo{year}{2020}\natexlab{}.
\newblock \showarticletitle{Domain generalization via entropy regularization}.
\newblock \bibinfo{journal}{\emph{NeurIPS}}  \bibinfo{volume}{33}
  (\bibinfo{year}{2020}), \bibinfo{pages}{16096--16107}.
\newblock


\bibitem[Zhou et~al\mbox{.}(2023)]%
        {zhou2022domain}
\bibfield{author}{\bibinfo{person}{Kaiyang Zhou}, \bibinfo{person}{Ziwei Liu},
  \bibinfo{person}{Yu Qiao}, \bibinfo{person}{Tao Xiang}, {and}
  \bibinfo{person}{Chen~Change Loy}.} \bibinfo{year}{2023}\natexlab{}.
\newblock \showarticletitle{Domain Generalization: A Survey}.
\newblock \bibinfo{journal}{\emph{IEEE Transactions on Pattern Analysis and
  Machine Intelligence}} \bibinfo{volume}{45}, \bibinfo{number}{4}
  (\bibinfo{year}{2023}), \bibinfo{pages}{4396--4415}.
\newblock
\href{https://doi.org/10.1109/TPAMI.2022.3195549}{doi:\nolinkurl{10.1109/TPAMI.2022.3195549}}


\bibitem[Zhu et~al\mbox{.}(2024)]%
        {zhu2024advancing}
\bibfield{author}{\bibinfo{person}{Chunzheng Zhu}, \bibinfo{person}{Jianxin
  Lin}, \bibinfo{person}{Guanghua Tan}, \bibinfo{person}{Ningbo Zhu},
  \bibinfo{person}{Kenli Li}, \bibinfo{person}{Chunlian Wang}, {and}
  \bibinfo{person}{Shengli Li}.} \bibinfo{year}{2024}\natexlab{}.
\newblock \showarticletitle{Advancing Ultrasound Medical Continuous Learning
  with Task-Specific Generalization and Adaptability}. In
  \bibinfo{booktitle}{\emph{2024 IEEE International Conference on
  Bioinformatics and Biomedicine (BIBM)}}. \bibinfo{publisher}{IEEE},
  \bibinfo{address}{Lisbon, Portugal}, \bibinfo{pages}{3019--3025}.
\newblock
\href{https://doi.org/10.1109/BIBM62325.2024.10822568}{doi:\nolinkurl{10.1109/BIBM62325.2024.10822568}}


\bibitem[Zhu et~al\mbox{.}(2026)]%
        {zhu2026medeyes}
\bibfield{author}{\bibinfo{person}{Chunzheng Zhu}, \bibinfo{person}{Yangfang
  Lin}, \bibinfo{person}{Shen Chen}, \bibinfo{person}{Yijun Wang}, {and}
  \bibinfo{person}{Jianxin Lin}.} \bibinfo{year}{2026}\natexlab{}.
\newblock \showarticletitle{MedEyes: Learning Dynamic Visual Focus for Medical
  Progressive Diagnosis}.
\newblock \bibinfo{journal}{\emph{Proceedings of the AAAI Conference on
  Artificial Intelligence}} \bibinfo{volume}{40}, \bibinfo{number}{16}
  (\bibinfo{year}{2026}), \bibinfo{pages}{13916--13924}.
\newblock
\href{https://doi.org/10.1609/aaai.v40i16.38401}{doi:\nolinkurl{10.1609/aaai.v40i16.38401}}


\bibitem[Zhu et~al\mbox{.}(2025)]%
        {zhu2025pathology}
\bibfield{author}{\bibinfo{person}{Chunzheng Zhu}, \bibinfo{person}{Yangfang
  Lin}, \bibinfo{person}{Jialin Shao}, \bibinfo{person}{Jianxin Lin}, {and}
  \bibinfo{person}{Yijun Wang}.} \bibinfo{year}{2025}\natexlab{}.
\newblock \showarticletitle{Pathology-Aware Prototype Evolution via LLM-Driven
  Semantic Disambiguation for Multicenter Diabetic Retinopathy Diagnosis}. In
  \bibinfo{booktitle}{\emph{Proceedings of the 33rd ACM International
  Conference on Multimedia}}. \bibinfo{publisher}{ACM},
  \bibinfo{address}{Dublin, Ireland}, \bibinfo{pages}{9196--9205}.
\newblock
\href{https://doi.org/10.1145/3746027.3754562}{doi:\nolinkurl{10.1145/3746027.3754562}}


\end{thebibliography}

\end{document}